%% file: main.tex
\pdfoutput=1   
\documentclass{article}
\usepackage{iclr2027_conference,times}
\iclrfinalcopy

\usepackage{amsmath,amssymb}
\usepackage{booktabs}
\usepackage{graphicx}
\usepackage{array}
\usepackage{multirow}
\usepackage{xcolor}
\usepackage{microtype}
\usepackage{hyperref}
\usepackage{url}
\usepackage{tikz}
\usetikzlibrary{positioning,arrows.meta,fit,backgrounds,calc}

\input{results_macros}          

\newcommand{\merit}{\textsc{Merit}}
\newcommand{\cUnm}{\textsc{unmatched}}
\newcommand{\cMat}{\textsc{matched}}
\newcommand{\cSame}{\textsc{same-talker}}
\newcommand{\cSameM}{\textsc{same-talker matched}}

\title{\textsc{Merit}: Making Every Modality\\
       Earn Its Contribution in\\
       Multimodal Attention Decoding}

\author{Ali Alavi \& Donald S.~Williamson\\
Department of Computer Science and Engineering\\
The Ohio State University\\
Columbus, OH 43210, USA\\
\texttt{\{alavibajestan.1,williamson.413\}@osu.edu}}

\hypersetup{pdftitle={MERIT: Making Every Modality Earn Its Contribution in Multimodal Attention Decoding},
            pdfauthor={Ali Alavi, Donald S. Williamson}}

\begin{document}
\maketitle
\lhead{}   

\input{sections/00_abstract}
\input{sections/01_intro}
\input{sections/07_related}
\input{sections/02_setting}
\input{sections/03_failures}
\input{sections/04_method}
\input{sections/05_setup}
\input{sections/06_results}
\input{sections/08_limitations}
\input{sections/09_statements}

\bibliographystyle{iclr2027_conference}
\bibliography{refs}

\appendix
\input{sections/A_failures}
\input{sections/A_related}
\input{sections/A_appendix}

\end{document}

%% file: results_macros.tex
\makeatletter
\providecommand{\defres}[2]{\expandafter\gdef\csname res@#1\endcsname{#2}}
\providecommand{\R}[1]{\ifcsname res@#1\endcsname\csname res@#1\endcsname
  \else\textcolor{red}{\texttt{??}}\fi}
\makeatother
\defres{extdtuloso/eeg_acausal/loso/contrib}{+0.2060}
\defres{extdtuloso/eeg_causal/loso/contrib}{+0.2006}
\defres{extdtuloso/eeg_centred/loso/acc}{0.7269}
\defres{extdtuloso/eeg_centred/loso/contrib}{+0.2264}
\defres{extdtuloso/eeg_centred/loso/pos}{18/18}
\defres{extdtuloso5/eeg_centred/loso/acc}{0.6635}
\defres{extdtuloso5/eeg_centred/loso/nfolds}{18}
\defres{extdtuwithin/eeg_acausal/within/contrib}{+0.2499}
\defres{extdtuwithin/eeg_causal/within/contrib}{+0.2560}
\defres{extdtuwithin/eeg_centred/within/acc}{0.7590}
\defres{extdtuwithin/eeg_centred/within/contrib}{+0.2584}
\defres{extdtuwithin/eeg_centred/within/pos}{5/5}
\defres{extdtuwithin5/eeg_centred/within/acc}{0.6909}
\defres{extdtuwithinst/eeg_acausal/within/contrib}{+0.3145}
\defres{extdtuwithinst/eeg_causal/within/contrib}{+0.3164}
\defres{extdtuwithinst/eeg_centred/within/acc}{0.8170}
\defres{extdtuwithinst/eeg_centred/within/contrib}{+0.3175}
\defres{extdtuwithinst/eeg_centred/within/pos}{5/5}
\defres{extkulloso/eeg_acausal/loso/contrib}{+0.1687}
\defres{extkulloso/eeg_causal/loso/contrib}{+0.1764}
\defres{extkulloso/eeg_centred/loso/acc}{0.7138}
\defres{extkulloso/eeg_centred/loso/contrib}{+0.1766}
\defres{extkulloso/eeg_centred/loso/pos}{16/16}
\defres{extkulloso5/eeg_centred/loso/acc}{0.6664}
\defres{extkulloso5/eeg_centred/loso/nfolds}{16}
\defres{extkulwithin/eeg_acausal/within/contrib}{+0.2074}
\defres{extkulwithin/eeg_causal/within/contrib}{+0.2104}
\defres{extkulwithin/eeg_centred/within/acc}{0.7252}
\defres{extkulwithin/eeg_centred/within/contrib}{+0.2141}
\defres{extkulwithin/eeg_centred/within/pos}{5/5}
\defres{extkulwithin5/eeg_centred/within/acc}{0.6763}
\defres{extkulwithinst/eeg_acausal/within/contrib}{+0.3080}
\defres{extkulwithinst/eeg_causal/within/contrib}{+0.3115}
\defres{extkulwithinst/eeg_centred/within/acc}{0.8149}
\defres{extkulwithinst/eeg_centred/within/contrib}{+0.3150}
\defres{extkulwithinst/eeg_centred/within/pos}{5/5}
\defres{extprior/dtu/prior}{0.494}
\defres{extprior/dtu/role}{male}
\defres{extprior/kul/nrole}{256}
\defres{extprior/kul/ntrials}{320}
\defres{extprior/kul/prior}{0.800}
\defres{extprior/kul/role}{track 1}
\defres{extprobe/dtu/w10/excess}{+0.0008}
\defres{extprobe/dtu/w10/nlisteners}{18}
\defres{extprobe/dtu/w10/nwin}{9{,}720}
\defres{extprobe/dtu/w10/probe}{0.5008}
\defres{extprobe/dtu/w2/excess}{+0.0010}
\defres{extprobe/dtu/w30/excess}{-0.0019}
\defres{extprobe/dtu/w5/excess}{+0.0001}
\defres{extprobe/kul/w10/excess}{+0.1947}
\defres{extprobe/kul/w10/nlisteners}{16}
\defres{extprobe/kul/w10/nwin}{9{,}564}
\defres{extprobe/kul/w10/probe}{0.6947}
\defres{extprobe/kul/w10/qmexcess}{+0.0974}
\defres{extprobe/kul/w2/excess}{+0.0945}
\defres{extprobe/kul/w2/qmexcess}{+0.0321}
\defres{extprobe/kul/w30/excess}{+0.0729}
\defres{extprobe/kul/w30/nwin}{4{,}608}
\defres{extprobe/kul/w30/qmexcess}{+0.0208}
\defres{extprobe/kul/w5/excess}{+0.1384}
\defres{extprobe/kul/w5/qmexcess}{+0.0401}
\defres{hinge/all_hinge_accsel/within/acc}{0.7087}
\defres{hinge/all_hinge_accsel/within/contrib}{+0.4486}
\defres{hinge/all_hinge_accsel/within/null}{0.2602}
\defres{hinge/all_hinge_accsel/within/shap/eeg}{+0.0882}
\defres{hinge/all_hinge_accsel/within/shap/fovea}{+0.2736}
\defres{hinge/all_hinge_accsel/within/shap/gaze}{+0.0442}
\defres{hinge/all_hinge_accsel/within/shap/imu}{+0.0156}
\defres{hinge/all_hinge_accsel/within/shap/video}{+0.0270}
\defres{hinge/all_hinge_credsel/within/acc}{0.7088}
\defres{hinge/all_hinge_credsel/within/contrib}{+0.4473}
\defres{hinge/all_hinge_credsel/within/null}{0.2614}
\defres{hinge/all_hinge_credsel/within/shap/eeg}{+0.0931}
\defres{hinge/all_hinge_credsel/within/shap/fovea}{+0.2653}
\defres{hinge/all_hinge_credsel/within/shap/gaze}{+0.0446}
\defres{hinge/all_hinge_credsel/within/shap/imu}{+0.0186}
\defres{hinge/all_hinge_credsel/within/shap/video}{+0.0258}
\defres{hinge/all_hinge_nodrop/within/acc}{0.7007}
\defres{hinge/all_hinge_nodrop/within/contrib}{+0.4458}
\defres{hinge/all_hinge_nodrop/within/null}{0.2549}
\defres{hinge/all_hinge_nodrop/within/shap/eeg}{+0.1236}
\defres{hinge/all_hinge_nodrop/within/shap/fovea}{+0.2376}
\defres{hinge/all_hinge_nodrop/within/shap/gaze}{+0.0393}
\defres{hinge/all_hinge_nodrop/within/shap/imu}{+0.0156}
\defres{hinge/all_hinge_nodrop/within/shap/video}{+0.0297}
\defres{hinge/all_hinge_uniform/within/acc}{0.7283}
\defres{hinge/all_hinge_uniform/within/contrib}{+0.4688}
\defres{hinge/all_hinge_uniform/within/null}{0.2594}
\defres{hinge/all_hinge_uniform/within/shap/eeg}{+0.0785}
\defres{hinge/all_hinge_uniform/within/shap/fovea}{+0.3135}
\defres{hinge/all_hinge_uniform/within/shap/gaze}{+0.0362}
\defres{hinge/all_hinge_uniform/within/shap/imu}{+0.0124}
\defres{hinge/all_hinge_uniform/within/shap/video}{+0.0282}
\defres{hinge/all_nohinge_accsel/within/acc}{0.7154}
\defres{hinge/all_nohinge_accsel/within/contrib}{+0.4554}
\defres{hinge/all_nohinge_accsel/within/null}{0.2600}
\defres{hinge/all_nohinge_accsel/within/shap/eeg}{+0.1039}
\defres{hinge/all_nohinge_accsel/within/shap/fovea}{+0.2954}
\defres{hinge/all_nohinge_accsel/within/shap/gaze}{+0.0296}
\defres{hinge/all_nohinge_accsel/within/shap/imu}{+0.0087}
\defres{hinge/all_nohinge_accsel/within/shap/video}{+0.0178}
\defres{hinge/all_nohinge_credsel/within/acc}{0.7250}
\defres{hinge/all_nohinge_credsel/within/contrib}{+0.4611}
\defres{hinge/all_nohinge_credsel/within/null}{0.2639}
\defres{hinge/all_nohinge_credsel/within/shap/eeg}{+0.1182}
\defres{hinge/all_nohinge_credsel/within/shap/fovea}{+0.2816}
\defres{hinge/all_nohinge_credsel/within/shap/gaze}{+0.0307}
\defres{hinge/all_nohinge_credsel/within/shap/imu}{+0.0109}
\defres{hinge/all_nohinge_credsel/within/shap/video}{+0.0197}
\defres{hinge/all_nohinge_nodrop/within/acc}{0.7226}
\defres{hinge/all_nohinge_nodrop/within/contrib}{+0.4684}
\defres{hinge/all_nohinge_nodrop/within/null}{0.2542}
\defres{hinge/all_nohinge_nodrop/within/shap/eeg}{+0.1207}
\defres{hinge/all_nohinge_nodrop/within/shap/fovea}{+0.2999}
\defres{hinge/all_nohinge_nodrop/within/shap/gaze}{+0.0254}
\defres{hinge/all_nohinge_nodrop/within/shap/imu}{+0.0064}
\defres{hinge/all_nohinge_nodrop/within/shap/video}{+0.0161}
\defres{ladder_qm/L0_published/within/acc}{0.2833}
\defres{ladder_raw/L0_published/within/acc}{0.4991}
\defres{ladder_raw/L0_published/within/collapse}{0.9138}
\defres{ladder_raw/L0_published/within/contrib}{+0.0000}
\defres{ladder_raw/L0_published/within/flip}{0.0000}
\defres{ladder_raw/L1_encoder/within/acc}{0.5800}
\defres{ladder_raw/L1_encoder/within/contrib}{+0.0000}
\defres{lag_mm/acausal/within/acc}{0.6486}
\defres{lag_mm/acausal/within/contrib}{+0.1478}
\defres{lag_mm/acausal/within/null}{0.5008}
\defres{lag_mm/causal/within/acc}{0.8085}
\defres{lag_mm/causal/within/contrib}{+0.2959}
\defres{lag_mm/causal/within/null}{0.5126}
\defres{lag_mm/centred/within/acc}{0.6896}
\defres{lag_mm/centred/within/contrib}{+0.1905}
\defres{lag_mm/centred/within/pos}{5/5}
\defres{lag_qm/acausal/within/acc}{0.4171}
\defres{lag_qm/acausal/within/contrib}{+0.0967}
\defres{lag_qm/acausal/within/null}{0.3204}
\defres{lag_qm/causal/within/acc}{0.4990}
\defres{lag_qm/causal/within/contrib}{+0.2240}
\defres{lag_qm/causal/within/null}{0.2750}
\defres{loso/all/loso/acc}{0.6706}
\defres{loso/all/loso/contrib}{+0.4038}
\defres{loso/all/loso/null}{0.2668}
\defres{loso/all/loso/pos}{16/16}
\defres{loso/all/loso/p}{0.0476}
\defres{loso/behaviour/loso/acc}{0.6300}
\defres{loso/behaviour/loso/contrib}{+0.3623}
\defres{loso/behaviour/loso/null}{0.2677}
\defres{loso/behaviour/loso/pos}{16/16}
\defres{loso/behaviour/loso/p}{0.0476}
\defres{loso/eeg/loso/acc}{0.4831}
\defres{loso/eeg/loso/contrib}{+0.1701}
\defres{loso/eeg/loso/null}{0.3130}
\defres{loso/eeg/loso/pos}{15/16}
\defres{loso/eeg/loso/p}{0.0863}
\defres{loso/eeg_flow/loso/acc}{0.5013}
\defres{loso/eeg_flow/loso/contrib}{+0.2245}
\defres{loso/eeg_flow/loso/null}{0.2767}
\defres{loso/eeg_flow/loso/pos}{16/16}
\defres{loso/eeg_flow/loso/p}{0.0506}
\defres{loso/eeg_fovea/loso/acc}{0.6700}
\defres{loso/eeg_fovea/loso/contrib}{+0.4051}
\defres{loso/eeg_fovea/loso/null}{0.2649}
\defres{loso/eeg_fovea/loso/pos}{16/16}
\defres{loso/eeg_fovea/loso/p}{0.0476}
\defres{loso/eeg_gaze/loso/acc}{0.4844}
\defres{loso/eeg_gaze/loso/contrib}{+0.2084}
\defres{loso/eeg_gaze/loso/null}{0.2760}
\defres{loso/eeg_gaze/loso/pos}{16/16}
\defres{loso/eeg_gaze/loso/p}{0.0506}
\defres{loso/eeg_imu/loso/acc}{0.4944}
\defres{loso/eeg_imu/loso/contrib}{+0.2162}
\defres{loso/eeg_imu/loso/null}{0.2782}
\defres{loso/eeg_imu/loso/pos}{16/16}
\defres{loso/eeg_imu/loso/p}{0.0536}
\defres{mm2/all/within/acc}{0.6857}
\defres{mm2/all/within/contrib}{+0.1869}
\defres{mm2/all/within/null}{0.4988}
\defres{mm2/eeg/within/acc}{0.6981}
\defres{mm2/eeg/within/contrib}{+0.1986}
\defres{mm2/eeg/within/null}{0.4995}
\defres{mm3/all/within/acc}{0.5578}
\defres{mm3/all/within/contrib}{+0.2257}
\defres{mm3/all/within/null}{0.3321}
\defres{mm3/eeg/within/acc}{0.5474}
\defres{mm3/eeg/within/contrib}{+0.2156}
\defres{mm3/eeg/within/null}{0.3318}
\defres{probe/qmatch/K4/excess}{+0.0103}
\defres{probe/qmatch/K4}{0.2603}
\defres{probe/raw/K4/chance}{0.2500}
\defres{probe/raw/K4/excess}{+0.3081}
\defres{probe/raw/K4}{0.5581}
\defres{probe/shifted/K2/excess}{+0.0131}
\defres{probe/shifted/K2}{0.5131}
\defres{probe/shifted/K3/excess}{+0.0425}
\defres{probe/shifted/K3}{0.3758}
\defres{probe/shifted_qm/K2/excess}{+0.0134}
\defres{probe/shifted_qm/K2}{0.5134}
\defres{probe/shifted_qm/K3/excess}{+0.0231}
\defres{probe/shifted_qm/K3}{0.3564}
\defres{sep/envelopecrestdB/d}{-1.22}
\defres{streams/all/within/acc}{0.7101}
\defres{streams/all/within/contrib}{+0.4504}
\defres{streams/all/within/null}{0.2597}
\defres{streams/all/within/pos}{5/5}
\defres{streams/all/within/p}{0.0476}
\defres{streams/behaviour/within/acc}{0.6552}
\defres{streams/behaviour/within/contrib}{+0.4010}
\defres{streams/behaviour/within/null}{0.2541}
\defres{streams/behaviour/within/pos}{5/5}
\defres{streams/behaviour/within/p}{0.0476}
\defres{streams/eeg/within/acc}{0.4788}
\defres{streams/eeg/within/contrib}{+0.1906}
\defres{streams/eeg/within/null}{0.2882}
\defres{streams/eeg/within/pos}{5/5}
\defres{streams/eeg/within/p}{0.0476}
\defres{streams/eeg_flow/within/acc}{0.5325}
\defres{streams/eeg_flow/within/contrib}{+0.2471}
\defres{streams/eeg_flow/within/null}{0.2854}
\defres{streams/eeg_flow/within/pos}{5/5}
\defres{streams/eeg_flow/within/p}{0.0476}
\defres{streams/eeg_fovea/within/acc}{0.6909}
\defres{streams/eeg_fovea/within/contrib}{+0.4358}
\defres{streams/eeg_fovea/within/null}{0.2550}
\defres{streams/eeg_fovea/within/pos}{5/5}
\defres{streams/eeg_fovea/within/p}{0.0476}
\defres{streams/eeg_gaze/within/acc}{0.5295}
\defres{streams/eeg_gaze/within/contrib}{+0.2685}
\defres{streams/eeg_gaze/within/null}{0.2610}
\defres{streams/eeg_gaze/within/pos}{5/5}
\defres{streams/eeg_gaze/within/p}{0.0476}
\defres{streams/eeg_imu/within/acc}{0.4995}
\defres{streams/eeg_imu/within/contrib}{+0.2405}
\defres{streams/eeg_imu/within/null}{0.2589}
\defres{streams/eeg_imu/within/pos}{5/5}
\defres{streams/eeg_imu/within/p}{0.0476}
\defres{v2bestloso/eeg/loso/acc}{0.6000}
\defres{v2bestloso/eeg/loso/contrib}{+0.3132}
\defres{v2hinge2/H1_hinge_drop/within/acc}{0.7884}
\defres{v2hinge2/H1_hinge_drop/within/contrib}{+0.5111}
\defres{v2hinge2/H1_hinge_drop/within/null}{0.2774}
\defres{v2hinge2/H1_hinge_drop/within/shap/eeg}{+0.1600}
\defres{v2hinge2/H1_hinge_drop/within/shap/fovea}{+0.2758}
\defres{v2hinge2/H1_hinge_drop/within/shap/gaze}{+0.0362}
\defres{v2hinge2/H1_hinge_drop/within/shap/imu}{+0.0144}
\defres{v2hinge2/H1_hinge_drop/within/shap/video}{+0.0247}
\defres{v2hinge2/H2_nohinge_drop/within/acc}{0.8091}
\defres{v2hinge2/H2_nohinge_drop/within/contrib}{+0.5298}
\defres{v2hinge2/H2_nohinge_drop/within/null}{0.2793}
\defres{v2hinge2/H2_nohinge_drop/within/shap/eeg}{+0.2229}
\defres{v2hinge2/H2_nohinge_drop/within/shap/fovea}{+0.2656}
\defres{v2hinge2/H2_nohinge_drop/within/shap/gaze}{+0.0206}
\defres{v2hinge2/H2_nohinge_drop/within/shap/imu}{+0.0079}
\defres{v2hinge2/H2_nohinge_drop/within/shap/video}{+0.0128}
\defres{v2hinge2/H3_hinge_nodrop/within/acc}{0.8242}
\defres{v2hinge2/H3_hinge_nodrop/within/contrib}{+0.5557}
\defres{v2hinge2/H3_hinge_nodrop/within/null}{0.2685}
\defres{v2hinge2/H3_hinge_nodrop/within/shap/eeg}{+0.2169}
\defres{v2hinge2/H3_hinge_nodrop/within/shap/fovea}{+0.2679}
\defres{v2hinge2/H3_hinge_nodrop/within/shap/gaze}{+0.0340}
\defres{v2hinge2/H3_hinge_nodrop/within/shap/imu}{+0.0074}
\defres{v2hinge2/H3_hinge_nodrop/within/shap/video}{+0.0295}
\defres{v2hinge2/H4_nohinge_nodrop/within/acc2}{0.82}
\defres{v2hinge2/H4_nohinge_nodrop/within/acc}{0.8173}
\defres{v2hinge2/H4_nohinge_nodrop/within/contrib}{+0.5439}
\defres{v2hinge2/H4_nohinge_nodrop/within/null}{0.2733}
\defres{v2hinge2/H4_nohinge_nodrop/within/shap/eeg}{+0.2217}
\defres{v2hinge2/H4_nohinge_nodrop/within/shap/fovea}{+0.2866}
\defres{v2hinge2/H4_nohinge_nodrop/within/shap/gaze}{+0.0194}
\defres{v2hinge2/H4_nohinge_nodrop/within/shap/imu}{+0.0039}
\defres{v2hinge2/H4_nohinge_nodrop/within/shap/video}{+0.0123}
\defres{v2hinge2loso/H1_hinge_drop/loso/acc}{0.7469}
\defres{v2hinge2loso/H1_hinge_drop/loso/pos}{16/16}
\defres{v2hinge2loso/H1_hinge_drop/loso/shap/eeg}{+0.1149}
\defres{v2hinge2loso/H2_nohinge_drop/loso/acc}{0.7625}
\defres{v2hinge2loso/H2_nohinge_drop/loso/pos}{16/16}
\defres{v2hinge2loso/H2_nohinge_drop/loso/shap/eeg}{+0.1560}
\defres{v2hinge2loso/H3_hinge_nodrop/loso/acc}{0.7219}
\defres{v2hinge2loso/H3_hinge_nodrop/loso/pos}{16/16}
\defres{v2hinge2loso/H3_hinge_nodrop/loso/shap/eeg}{+0.1479}
\defres{v2hinge2loso/H4_nohinge_nodrop/loso/acc2}{0.74}
\defres{v2hinge2loso/H4_nohinge_nodrop/loso/acc}{0.7438}
\defres{v2hinge2loso/H4_nohinge_nodrop/loso/contrib}{+0.4737}
\defres{v2hinge2loso/H4_nohinge_nodrop/loso/pos}{16/16}
\defres{v2hinge2loso/H4_nohinge_nodrop/loso/shap/eeg}{+0.1762}
\defres{v2ladder/V4_subjadapt/within/acc}{0.6709}
\defres{v2onset/V1_crops/within/acc}{0.4956}
\defres{v2onset/V1_crops/within/contrib}{+0.1974}
\defres{v2onset/V1_crops/within/null}{0.2982}
\defres{v2onset/V2_trial/within/acc}{0.6676}
\defres{v2onset/V2_trial/within/contrib}{+0.2866}
\defres{v2onset/V2_trial/within/null}{0.3810}
\defres{v2onset/V3_causal/within/acc}{0.6502}
\defres{v2onset/V3_causal/within/contrib}{+0.3271}
\defres{v2onset/V3_causal/within/null}{0.3231}
\defres{v2onset/V3b_acausal/within/acc}{0.6145}
\defres{v2onset/V3b_acausal/within/contrib}{+0.2127}
\defres{v2onset/V3b_acausal/within/null}{0.4017}
\defres{v2onset/V4_subjadapt/within/acc}{0.6915}
\defres{v2onset/V4_subjadapt/within/contrib}{+0.3971}
\defres{v2onset/V4_subjadapt/within/null}{0.2944}
\defres{w2/all/within/acc}{0.7291}
\defres{w2/all/within/contrib}{+0.4682}
\defres{w2/all/within/null}{0.2608}
\defres{w2/all/within/p}{0.0050}
\defres{w2/behaviour/within/acc}{0.7174}
\defres{w2/behaviour/within/contrib}{+0.4599}
\defres{w2/behaviour/within/null}{0.2575}
\defres{w2/behaviour/within/p}{0.0050}
\defres{w2/eeg/within/acc}{0.3281}
\defres{w2/eeg/within/contrib}{+0.0263}
\defres{w2/eeg/within/null}{0.3018}
\defres{w2/eeg/within/p}{0.0647}
\defres{w2/eeg_flow/within/acc}{0.4509}
\defres{w2/eeg_flow/within/contrib}{+0.1569}
\defres{w2/eeg_flow/within/null}{0.2939}
\defres{w2/eeg_flow/within/p}{0.0050}
\defres{w2/eeg_fovea/within/acc}{0.7251}
\defres{w2/eeg_fovea/within/contrib}{+0.4606}
\defres{w2/eeg_fovea/within/null}{0.2645}
\defres{w2/eeg_fovea/within/p}{0.0050}
\defres{w2/eeg_gaze/within/acc}{0.4167}
\defres{w2/eeg_gaze/within/contrib}{+0.1365}
\defres{w2/eeg_gaze/within/null}{0.2803}
\defres{w2/eeg_gaze/within/p}{0.0050}
\defres{w2/eeg_imu/within/acc}{0.4134}
\defres{w2/eeg_imu/within/contrib}{+0.1182}
\defres{w2/eeg_imu/within/null}{0.2952}
\defres{w2/eeg_imu/within/p}{0.0050}
\defres{w2/flow/within/acc}{0.4109}
\defres{w2/flow/within/contrib}{+0.1632}
\defres{w2/flow/within/null}{0.2477}
\defres{w2/flow/within/p}{0.0050}
\defres{w2/fovea/within/acc}{0.7130}
\defres{w2/fovea/within/contrib}{+0.4570}
\defres{w2/fovea/within/null}{0.2560}
\defres{w2/fovea/within/p}{0.0050}
\defres{w2/gaze/within/acc}{0.3448}
\defres{w2/gaze/within/contrib}{+0.1144}
\defres{w2/gaze/within/null}{0.2304}
\defres{w2/gaze/within/p}{0.0050}
\defres{w2/imu/within/acc}{0.3541}
\defres{w2/imu/within/contrib}{+0.1151}
\defres{w2/imu/within/null}{0.2389}
\defres{w2/imu/within/p}{0.0050}
\defres{w5/all/within/acc}{0.7189}
\defres{w5/all/within/contrib}{+0.4547}
\defres{w5/all/within/null}{0.2642}
\defres{w5/all/within/p}{0.0050}
\defres{w5/behaviour/within/acc}{0.6857}
\defres{w5/behaviour/within/contrib}{+0.4282}
\defres{w5/behaviour/within/null}{0.2575}
\defres{w5/behaviour/within/p}{0.0050}
\defres{w5/eeg/within/acc}{0.3954}
\defres{w5/eeg/within/contrib}{+0.0853}
\defres{w5/eeg/within/null}{0.3101}
\defres{w5/eeg/within/p}{0.0418}
\defres{w5/eeg_flow/within/acc}{0.4994}
\defres{w5/eeg_flow/within/contrib}{+0.1992}
\defres{w5/eeg_flow/within/null}{0.3003}
\defres{w5/eeg_flow/within/p}{0.0050}
\defres{w5/eeg_fovea/within/acc}{0.7198}
\defres{w5/eeg_fovea/within/contrib}{+0.4450}
\defres{w5/eeg_fovea/within/null}{0.2748}
\defres{w5/eeg_fovea/within/p}{0.0050}
\defres{w5/eeg_gaze/within/acc}{0.4845}
\defres{w5/eeg_gaze/within/contrib}{+0.1918}
\defres{w5/eeg_gaze/within/null}{0.2927}
\defres{w5/eeg_gaze/within/p}{0.0050}
\defres{w5/eeg_imu/within/acc}{0.4914}
\defres{w5/eeg_imu/within/contrib}{+0.1716}
\defres{w5/eeg_imu/within/null}{0.3198}
\defres{w5/eeg_imu/within/p}{0.0050}
\defres{w5/flow/within/acc}{0.4126}
\defres{w5/flow/within/contrib}{+0.1668}
\defres{w5/flow/within/null}{0.2458}
\defres{w5/flow/within/p}{0.0050}
\defres{w5/fovea/within/acc}{0.7096}
\defres{w5/fovea/within/contrib}{+0.4530}
\defres{w5/fovea/within/null}{0.2566}
\defres{w5/fovea/within/p}{0.0050}
\defres{w5/gaze/within/acc}{0.3349}
\defres{w5/gaze/within/contrib}{+0.1113}
\defres{w5/gaze/within/null}{0.2236}
\defres{w5/gaze/within/p}{0.0050}
\defres{w5/imu/within/acc}{0.3800}
\defres{w5/imu/within/contrib}{+0.1345}
\defres{w5/imu/within/null}{0.2456}
\defres{w5/imu/within/p}{0.0050}

%% file: sections/00_abstract.tex
\begin{abstract}
More than $1.5$ billion people live with hearing loss, and when several people
talk at once, a hearing aid amplifies every voice alike. Auditory attention
decoding (AAD) aims to solve this: it reads a listener's brain signals, finds the
talker they are attending to, and lets the device amplify that talker. A real
device will not see the brain alone. Glasses and earbuds can also track the eyes,
the head and the scene, and these signals help, because people tend to look at
the person they listen to. But they show where a listener is looking, not whom
they are listening to. A decoder that leans on them works when the two agree and
fails when they differ --- when a listener cannot move their eyes or head, or
listens to someone they are not looking at --- which is exactly when a brain-based
device is needed. A multimodal decoder must therefore use these signals without
letting them replace the electroencephalogram (EEG), and accuracy alone cannot
show whether it does. We
present \merit{}, a decoder and an evaluation method built around one question:
how much of the accuracy does each input earn? \merit{} shuffles each input
stream across test examples, measures how much accuracy drops, and splits the
result exactly across the streams with Shapley values. It also blocks two
shortcuts by design: the candidate audio is checked before training so that it
cannot reveal the answer, and the scoring function cannot rise above chance if it
ignores the brain signal. On a public dataset with EEG, eye tracking, head motion
and scene video, \merit{} improves on the published benchmark, reaching
\R{v2hinge2/H4_nohinge_nodrop/within/acc2} four-way accuracy within listeners and
\R{v2hinge2loso/H4_nohinge_nodrop/loso/acc2} on unseen listeners, where chance is
$0.25$. The measurement also shows what accuracy hides: adding the behavioural
streams raises accuracy but almost halves the brain's share of it. On two other
public datasets, the same tests find a label imbalance in a widely used benchmark
and flag results that accuracy alone would report as brain decoding.
\end{abstract}

%% file: sections/01_intro.tex
\section{Introduction}
\label{sec:intro}

\paragraph{Why attention decoding matters.}
About $1.5$ billion people live with some hearing loss, and the number is
expected to reach $2.5$ billion by $2050$ \citep{who2021hearing}. The hardest
situation for them is the one \citet{cherry1953} called the cocktail party
problem: several people talking at once. A hearing aid can make sound louder, but
it cannot tell which voice matters, so it amplifies all of them, including the
ones the listener is trying to ignore. \emph{Auditory attention decoding} (AAD)
addresses this \citep{osullivan2015,geirnaert2021}. The brain's response follows
the speech a listener attends to \citep{ding2012,lalor2009}, so a decoder can
read that response from the electroencephalogram (EEG), find the attended
talker, and steer a hearing device toward them \citep{pan2024neuroheed}.

\paragraph{Why several streams matter.}
Such a device will not see the brain alone. Glasses and earbuds already carry
motion sensors, and eye trackers and cameras are being added to them. These
signals are useful. In everyday listening, people turn their head and eyes toward
the talker they follow \citep{hendrikse2019,best2023gaze,kidd2005}, and a hearing
aid steered by an eye tracker has already been built \citep{kidd2013visually}. A
realistic decoder should therefore combine these streams with the EEG, and doing
so raises accuracy (Section~\ref{sec:res-hinge}).

\paragraph{Why the EEG must still carry the decision.}
Gaze and head motion show where a listener is \emph{facing}, not whom they are
\emph{listening to}, and the two often differ. A driver listens to a passenger
while watching the road. At a dinner table, people listen to one person while
looking at another. Some listeners cannot move their eyes or head freely at all.
A decoder that leans on the behavioural streams does well when facing and
listening agree, and fails in exactly these cases. Those are the cases where a
brain-based device is needed: when facing and listening agree, an eye tracker is
enough. Whatever the EEG adds, it adds through the brain signal. A multimodal
decoder should therefore use the other streams as helpers, but it must not let
them replace the EEG. Accuracy cannot tell us whether this holds. A multimodal
model is reported with one number, and that number does not say which input
produced it. A decoder that silently decides from gaze and head motion and a
decoder that truly reads attention from the EEG can therefore report the same
accuracy, even though only the second keeps working when facing and listening
disagree. The risk is real: \citet{rotaru2024} found that decoders believed to
read spatial attention from EEG owed most of their accuracy to eye movements
captured in the recording.

The usual formulation is a matching task. The model sees a window of the
recording and $K$ candidate speech signals, one of which is the attended talker,
and it must pick that one; accuracy is compared with chance, $1/K$
\citep{accou2021,monesi2020,puffay2023}. We study it on MAESTRO
\citep{maestro2026}, a public dataset in which sixteen listeners each attend one
of four talkers speaking at once while EEG, eye tracking, head motion and scene
video are recorded together (Section~\ref{sec:setup}). We show that this task
can be solved without using the brain signal at all, in three ways
(Section~\ref{sec:failures}).
\textbf{(i) The audio gives the answer away.} Scaling each candidate to the same
loudness does not make the candidates look alike: statistics such as skew and
sparsity do not change with scale. On this dataset they differ between attended
and ignored talkers, and a simple classifier that sees only the audio reaches
\R{probe/raw/K4} four-way accuracy, against a chance of \R{probe/raw/K4/chance}.
\textbf{(ii) The standard scoring function gives the model a path to the leak.}
Problem (i) says the answer is \emph{present} in the audio; this one says the
standard architecture can \emph{reach} it without the brain. If the brain
encoder outputs the same vector at every time step --- the collapse an encoder
falls into when its input stops helping --- the usual score reduces to a
classifier on the audio alone, and training often finds this solution because it
is easier than the intended one.
\textbf{(iii) Easy streams crowd out the brain.} When several streams are
available, training leans on the easiest one
\citep{wang2020gradblend,huang2022competition,wu2022greedy,peng2022ogm}. Here
the easy streams measure where the listener is \emph{looking}, not what they are
\emph{listening to}, so accuracy rises while the evidence for the claim shrinks.

One test catches all three problems. We shuffle a recorded stream across the
test windows, so each window gets another window's recording but keeps its own
candidates and its own label, and we measure how much accuracy drops. We call the
drop the \textbf{contribution} of that stream. A decoder that does not use a
stream loses nothing when it is shuffled. This paper turns that test from an
occasional check into the centre of the method.

MAESTRO was released with a benchmark, and the benchmark already reports a
permutation test: it shuffles the EEG, gaze, head motion and video
\emph{together}, and its contributions are positive. That shows the listener's signals are used. It cannot show \emph{which}
signal is used, and problem (iii) sits exactly there, because a joint test gives
its largest value when a strong behavioural stream has pushed the brain aside.
\merit{} shuffles each stream \emph{separately} and splits the total exactly
across streams. It also decodes better than the benchmark:
\R{v2hinge2/H4_nohinge_nodrop/within/acc2} four-way accuracy within listeners and
\R{v2hinge2loso/H4_nohinge_nodrop/loso/acc2} on unseen listeners, against the
benchmark's $0.698$ and $0.674$ (Section~\ref{sec:res-full}).

\paragraph{Contributions.}
\textbf{(1) A per-stream measure of credit} (Section~\ref{sec:shapley}). We
compute each stream's Shapley value over all $2^M$ subsets of the $M$ recorded
streams ($M\!=\!5$ here), keeping the streams in a subset intact and shuffling
the rest.
The values add up exactly to the total contribution, and they are cheap enough
to report for every run.
\textbf{(2) A decoder in which the shortcuts are blocked by design}
(Section~\ref{sec:method}). A time-centred correlation score gives every
candidate the same score when the brain encoder is constant, so that shortcut
lands exactly at chance. The classifier heads that read gaze and head motion
receive no audio, so they cannot use audio cues. The candidate audio is tested before training.
\textbf{(3) Choosing checkpoints by contribution, not accuracy}, which trades a
little accuracy for a larger share of credit on the brain.
\textbf{(4) Findings.} Adding behavioural streams cuts the brain's share of the
credit almost in half; at two-second windows the EEG earns almost none of it.
On two other public datasets, the same tests find a label imbalance in a widely
used benchmark and show that high accuracy on both cannot be confirmed as brain
activity. We also tested adding the permutation test to the training loss
(Section~\ref{sec:hinges}); it does not increase the brain's share, and we report
this in full.

%% file: sections/07_related.tex
\section{Related work}
\label{sec:related}

\paragraph{Decoding auditory attention.}
Brain activity follows the speech a listener attends to \citep{ding2012,lalor2009}.
The oldest decoders are linear: they rebuild the speech envelope from the EEG and
compare the reconstruction with each talker \citep{osullivan2015}. They are
simple and well understood, but they need windows of tens of seconds to decide
reliably \citep{geirnaert2021}, too slow for a hearing device. Deep models
shorten the window in two ways. \emph{Match-mismatch} models learn directly
whether a window of EEG goes with a window of speech
\citep{accou2021,monesi2020,puffay2023}; our decoder belongs to this family.
\emph{Direction} decoders drop the audio and classify which side the listener
attends
\citep{vandecappelle2021,yan2024darnet,fan2025listennet,ni2024dbpnet,xu2024densenet,zhang2025swim}.
Both lines share a weakness: the reported accuracy often measures something other
than attention. \citet{rotaru2024} showed that direction decoding on several
public datasets relies largely on eye movements captured in the EEG rather than
on attention, and when within-trial leakage between training and test is removed,
recent direction decoders lose $17$ to $45$ points and several fall to chance
\citep{alavi2026neurotoken} (Table~\ref{tab:published}). We bring the concern of
\citet{rotaru2024} to the matching task and to several recorded streams at once,
and make the check part of training and model choice rather than an
after-the-fact test.

\paragraph{Datasets.}
The public AAD corpora record EEG and little else, and their designs deliberately
hold the listener still, which is good for isolating the brain response and
rules out the multimodal question: DTU uses earphones that simulate a room
\citep{fuglsang2018dtu}, KU~Leuven a quiet booth \citep{biesmans2017kul,das2019kuldata},
and a recent four-speaker dataset a chin rest \citep{yan2024fourspeaker}. A
stream that was never recorded, from a listener who was not free to move, cannot
compete with the EEG for credit. KU~Leuven has a further limitation we measure in
Appendix~\ref{app:external}: one of its two talkers is attended in
\R{extprior/kul/nrole} of \R{extprior/kul/ntrials} trials, so the audio alone
carries part of the answer. AV-GC-AAD \citep{geirnaert2024avgc} records gaze and
decouples it from attention on purpose --- the right tool for asking
\emph{whether} a decoder reads gaze instead of the brain, but its other streams
are designed to carry no information, so it cannot say how credit is shared among
streams that do. MAESTRO \citep{maestro2026} records EEG, eye tracking, head
motion and scene video from a listener who moves freely in a room of
loudspeakers, which is why we work on it; its own limitation, an audio confound
in the usual candidate construction, is the subject of
Section~\ref{sec:failures}.

\paragraph{Shortcut learning and multimodal balance.}
The problem is a case of shortcut learning
\citep{geirhos2020,torralba2011,gururangan2018}. The usual fixes rebalance the
data \citep{goyal2017vqa} or train against a shortcut-only branch
\citep{cadene2019rubi}; we add a scoring function in which the shortcut is
impossible. Work on balancing modalities
\citep{wang2020gradblend,huang2022competition,wu2022greedy,peng2022ogm} evens out
how fast each stream learns during training; we measure each stream's effect on
the decision at test time instead. The measurement is permutation importance
\citep{breiman2001,fisher2019} and the Shapley value
\citep{shapley1953,lundberg2017,covert2020} applied to whole streams; what is new
is using it to choose models and to report every result.
Appendix~\ref{app:related} gives a longer discussion.

%% file: sections/02_setting.tex
\section{Candidate sets and splits}
\label{sec:setting}

\paragraph{Candidate sets.}
\label{sec:constructions}
In the matching task the decoder is shown $K$ candidate speech signals and must
pick the one the listener attended. The candidates are inputs, not labels: the
decoder sees them, so anything in them that marks the correct one is a shortcut
that solves the task without the recording. How the candidates are built
therefore decides what the task measures, and we name each set by how it is
built. \cUnm{}: the four talkers of the trial, each standardised to zero mean
and unit variance. This is the usual choice in the literature. Standardising
removes loudness, but the four candidates are still four different recordings,
and any property beyond the mean and the variance --- skew, sparsity, the balance
between frequency bands --- can still tell them apart; on this dataset it does
(Section~\ref{sec:failures}). \cMat{}: the same four talkers after
\emph{quantile matching}. We sort each candidate's values, average the sorted
values across the four candidates, and give each candidate these averages back in
its own time order. All four then contain exactly the same set of values, so no
statistic of the values --- loudness, skew, sparsity or any other --- can tell
them apart. Each keeps its own timing, and timing is what a brain response
follows. This is the most we can remove from the audio without removing the
signal the task is about. \cSame{}: the wrong candidates are the \emph{attended
talker's own} speech, taken from other, non-overlapping parts of the same trial.
All candidates then come from one voice and one loudspeaker, so neither the
audio nor the listener's orientation can say which is correct, and chance is
exact by construction \citep{accou2021}. Trials are long enough for $K\!=\!3$ and
$K\!=\!2$. \cSameM{} applies both steps. We use \cMat{} for the four-way task
and \cSameM{} as the strictest check that the EEG follows the speech.

\paragraph{Splits.}
\emph{Within-listener}, five folds: trials are split by their audio content, so no
stimulus heard in training is heard again at test; every listener appears on both
sides. \emph{Leave-one-listener-out} (LOSO), sixteen folds: one listener is held
out, and $20\%$ of the audio content is also held out, so the test listener is new
in both identity and content. A validation set is always taken from the training
part, and each test set is scored once.

%% file: sections/03_failures.tex
\section{Three ways to be right for the wrong reason}
\label{sec:failures}

\paragraph{(i) The audio gives the answer away.}
\label{sec:conf}
The decoder sees the candidate audio, so anything in the audio that marks the
attended talker solves the task without the brain. Holding out listeners or
stimuli does not help, because the mark belongs to the \emph{role} of attended
talker, so it is present for every listener and every stimulus. Making all
candidates equally loud does not help either. Loudness is removed exactly by
standardising each candidate, but standardising only fixes the mean and the
variance. Every statistic that does not change with scale survives: skew,
kurtosis, sparsity and the balance between frequency bands. On this dataset the
attended and ignored talkers differ in exactly these. The peak-to-average ratio of
the envelope alone separates them with an effect size of
$d=\R{sep/envelopecrestdB/d}$. The dataset's authors trace this to the loudness
adjustment, which clipped the attended talker's peaks \citep{maestro2026}.

We therefore test each candidate set \emph{before} training, with an
\textbf{acceptance probe}: a logistic regression on eight statistics that do not
depend on scale, evaluated on held-out content. A clean candidate set must score
$1/K$. The \cUnm{} set scores \R{probe/raw/K4/excess} above chance.
\cMat{} brings this down to \R{probe/qmatch/K4} against $0.25$, and
same-talker candidates to \R{probe/shifted_qm/K2} against $0.50$
(Table~\ref{tab:probe}).

\paragraph{(ii) The standard scoring function has a shortcut.}
\label{sec:degenerate}
The standard score normalises the brain and audio embeddings at each time step,
multiplies them, averages over time and applies a linear layer
\citep{accou2021,monesi2020}. Suppose the brain encoder outputs the same unit
vector $\hat c$ at every step. This is not a hypothetical state: once the audio
already carries the answer, the easiest way for gradient descent to lower the
loss is to let the encoder ignore its noisy input and emit a constant, and we
observe exactly this collapse when the standard model is trained on this data
(Section~\ref{sec:res-ladder}). With a constant $\hat c$ the time average splits
apart and the score becomes $w^\top(\hat c \odot \frac1T\sum_t \hat a_{k,t}) + b$,
where $\hat a_{k,t}$ is the audio embedding of candidate $k$ at time step $t$,
$T$ is the number of time steps in the window, $\odot$ is the elementwise
product, and $w$ and $b$ are the weights and bias of the final linear layer.
This is a linear classifier on the time-averaged audio, with exactly the power
needed to read the audio mark of (i). So the architecture contains an audio-only
decoder, which it reaches by letting the brain encoder go constant. Problem (i)
makes the audio shortcut \emph{enough}; problem (ii) makes it \emph{reachable}.

\paragraph{(iii) Easy streams crowd out the brain.}
\label{sec:freeload}
When several streams are trained together, the model relies most on the easiest
one \citep{wang2020gradblend,huang2022competition,wu2022greedy,peng2022ogm}.
Here this changes what the result means. Gaze, head motion and scene video show
where the listener is \emph{facing}, which in a room of loudspeakers predicts the
attended talker well \citep{kidd2005,rotaru2024}. Only the EEG follows what the
talker is \emph{saying}. A fused decoder that quietly relies on gaze reports a
higher number while supporting a weaker claim.

As Section~\ref{sec:intro} argues, it also fails exactly for the listeners a brain-based device
is for: those who cannot, or choose not to, face the talker they listen to.

\paragraph{One test catches all three.}
\label{sec:perm}
Shuffling a recorded stream across test windows, while each window keeps its own
candidates and label, costs accuracy only if the decoder uses that stream. We
shuffle the recording, not the labels: shuffling the labels would also break the
link between the audio and the answer, and would hide the very shortcut we are
testing for (Appendix~\ref{app:perm}). Let
$v(T)$ be the accuracy when the streams in the set $T$ are left intact and the
others are shuffled. Then $v(\emptyset)$ is what \emph{this} model can get from the
audio alone, measured through its own audio pathway. The probe bounds what a
simple reader can take from the audio; the shuffled accuracy measures what this
decoder actually takes. Appendix~\ref{app:failures} gives all three problems in
more detail.

%% file: sections/04_method.tex
\section{\merit}
\label{sec:method}

\begin{figure}[t]
\centering
\definecolor{cIn}{HTML}{E8EFFB}\definecolor{cInB}{HTML}{3B6FB6}
\definecolor{cCo}{HTML}{E4EDFA}\definecolor{cCoB}{HTML}{3B6FB6}
\definecolor{cOr}{HTML}{E4F1EA}\definecolor{cOrB}{HTML}{3F8560}
\definecolor{cFz}{HTML}{F1EAF7}\definecolor{cFzB}{HTML}{8E5BA6}
\definecolor{cAd}{HTML}{FBEDE1}\definecolor{cAdB}{HTML}{BB6C24}
\resizebox{0.72\textwidth}{!}{%
\begin{tikzpicture}[font=\scriptsize,>=Stealth,x=1cm,y=1cm,
  nd/.style={draw,rounded corners=2pt,align=center,line width=0.6pt,inner sep=3pt},
  inp/.style={nd,fill=cIn,draw=cInB,minimum width=20mm,minimum height=5.6mm},
  enc/.style={nd,fill=white,draw=gray!60,minimum width=21mm,minimum height=5.6mm},
  frz/.style={nd,fill=cFz,draw=cFzB,minimum width=21mm,minimum height=5.6mm},
  fl/.style={->,line width=0.6pt,gray!50!black},
  fld/.style={->,line width=0.6pt,dashed,cAdB}]

  \node[inp] (eeg)  at (0, 0.00) {EEG $T\!\times\!32$};
  \node[inp] (gaze) at (0,-0.78) {gaze $T\!\times\!6$};
  \node[inp] (imu)  at (0,-1.56) {head $T\!\times\!6$};
  \node[inp] (flow) at (0,-2.34) {flow $T\!\times\!4$};
  \node[inp] (fov)  at (0,-3.12) {fovea $1024$};
  \node[inp] (aud)  at (0,-4.45) {$a_1\dots a_K$};

  \node[enc] (e1) at (3.1, 0.00) {dilated enc};
  \node[enc] (e2) at (3.1,-0.78) {dilated enc};
  \node[enc] (e3) at (3.1,-1.56) {dilated enc};
  \node[enc] (e4) at (3.1,-2.34) {dilated enc};
  \node[frz] (e5) at (3.1,-3.12) {\textbf{frozen} proj};
  \node[enc] (ea) at (3.1,-4.45) {shared audio enc};

  \node[nd,fill=cCo,draw=cCoB,minimum width=34mm,minimum height=7.5mm]
       (cpl) at (7.1, 0.00) {$s_k=\tau\,w^{\!\top}\mathrm{corr}_t(e,\hat a_k)$\\[-1pt]
                             \scriptsize centred over $t$ and over $k$};
  \node[nd,fill=cOr,draw=cOrB,minimum width=34mm,minimum height=11mm]
       (ori) at (7.1,-2.10) {speaker heads \emph{(no audio)}\\[-1pt]
                             \scriptsize gate $+$ fusion $\to$ 4 logits};
  \node[cCoB,font=\bfseries\tiny,anchor=south] at (7.1,0.55)
       {COUPLING BRANCH --- \emph{what} is attended};
  \node[cOrB,font=\bfseries\tiny,anchor=north] at (7.1,-2.72)
       {ORIENTATION BRANCH --- \emph{where} attention points};

  \node[nd,fill=cAd,draw=cAdB,minimum width=34mm,minimum height=5.6mm]
       (adv) at (7.1,-4.45) {audio-only adversary (GRL)};

  \node[nd,fill=gray!10,draw=gray!55,minimum width=17mm,minimum height=9mm]
       (sc) at (10.9,-1.05) {$K$ slot\\scores};
  \node[nd,fill=white,draw=cCoB,minimum width=30mm,minimum height=14mm]
       (loss) at (14.1,-1.05) {cross-entropy\\$+$ contrastive\\[1pt]\scriptsize $+$ EEG null hinge, anti-collapse};

  \foreach \a/\b in {eeg/e1,gaze/e2,imu/e3,flow/e4,fov/e5,aud/ea} \draw[fl] (\a)--(\b);
  \draw[fl] (e1) -- (cpl);
  \draw[fl] (ea.east) -- ++(0.35,0) |- (cpl.south west);
  \draw[fld] (ea) -- (adv);
  \draw[fl] (e1.east) -- ++(0.18,0) |- (ori.west);
  \foreach \e in {e2,e3,e4,e5} \draw[fl] (\e.east) -- ++(0.18,0) |- (ori.west);
  \draw[fl] (cpl.east) -- ++(0.4,0) |- (sc.north west);
  \draw[fl] (ori.east) -- ++(0.4,0) |- (sc.south west);
  \node[font=\tiny,anchor=west] at (9.05,-1.95) {gather by $\pi$};
  \draw[fl] (sc) -- (loss);
\end{tikzpicture}}
\caption{\textbf{\merit{}.} The coupling branch compares the EEG with each
candidate using a correlation over time with the mean removed. A brain encoder
that outputs the same vector at every step therefore gives every candidate a
score of zero, which is chance. The orientation branch gets \emph{no audio}, so
it cannot use audio cues. The model encodes once and scores separately, so a
trained model can be re-scored under any shuffle of any stream without encoding
again.}
\label{fig:arch}
\end{figure}
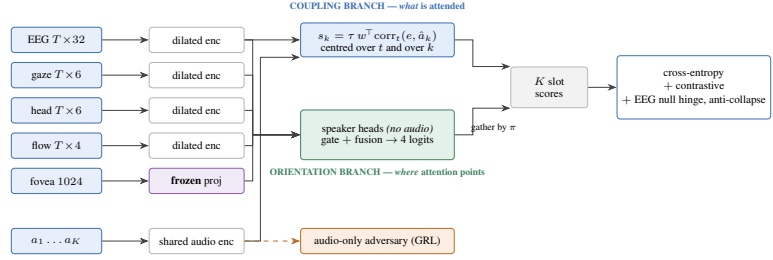

The name states the goal: every input should earn its share of the result.
\merit{} has two branches that answer two different questions. The
\textbf{coupling branch} asks whether the recording follows a candidate's speech
over time. Only EEG enters it, because only EEG has a timed relationship with a
speech signal. The \textbf{orientation branch} asks which loudspeaker the
listener was facing. It receives no audio, so it cannot use audio cues.
Figure~\ref{fig:arch} shows the whole decoder.

\paragraph{Encoders sized to the brain response.}
Each stream goes through five dilated temporal convolutions
\citep{yu2016dilated} with dilations $1,2,4,8,16$ and GroupNorm
\citep{wu2018groupnorm} after each layer. This gives a window of $63$ samples
($0.98$\,s at $64$\,Hz), about the time over which the cortical response to
speech unfolds \citep{lalor2009}. There is no activation on the last layer,
because a rectified output has no negative values and squeezes all the
correlations the score depends on toward one. Padding is symmetric, because the
brain responds $100$--$300$\,ms \emph{after} the sound, so an encoder that looks
only backward in time cannot see the response. Appendix~\ref{app:encoder}
details both choices. One-sided versions restrict the model to a chosen range of
delays (Section~\ref{sec:res-mm}).

\paragraph{A score that cannot use a constant brain signal.}
\label{sec:coupling}
For each of the $D\!=\!16$ embedding dimensions we compute the correlation over
time between the EEG embedding $e$ and the embedding $\hat a_k$ that the shared
audio encoder computes from the $k$-th candidate envelope. We
combine the $D$ values with a weight vector that has no bias, scale by a learned
temperature, and subtract the mean over candidates:
\begin{equation}
  \tilde e=\frac{e-\overline{e}}{\lVert e-\overline{e}\rVert_2},\qquad
  c_d(k)=\textstyle\sum_t \tilde e_{t,d}\,\tilde a_{k,t,d},\qquad
  s_k=\tau\, w^{\!\top} c(k),\qquad
  \mathrm{score}_k=s_k-\tfrac1K\textstyle\sum_j s_j .
  \label{eq:coupling}
\end{equation}
Removing the time average is the key step. If the encoder outputs the same
vector at every time step --- the state it falls into once it learns to ignore
its input --- then $e-\overline e$ is \emph{exactly zero}. Every $c_d(k)$ is
then zero, all $K$ scores are equal, and accuracy is exactly chance. The
shortcut of Section~\ref{sec:failures}(ii) is not discouraged by a penalty; it is
impossible. Two more properties follow. Correlation ignores scale, so a candidate
cannot win by being louder. Subtracting the mean over candidates removes anything
added equally to all of them, which is why $w$ has no bias.

\paragraph{Orientation heads with no audio, and an audio-only adversary.}
Gaze, head motion, scene motion and the foveal stream --- a frozen V-JEPA\,2
embedding of the patch of scene around the gaze point
(Section~\ref{sec:setup}) --- show where the listener is facing. They have no
timed relationship with speech, so asking them to match a speech envelope makes
no sense --- and doing so made them collapse in our first attempt. Instead, each
one feeds a classifier head that predicts the loudspeaker index from the mean
and standard deviation of its embedding over time. These classifier heads get
\textbf{no audio input}, so shuffling them drops accuracy to chance, and anything
above chance is earned by the recording. The EEG also gets such a classifier
head, because attention shifts brain activity toward one side. A learned, non-negative weight
per stream and a small fusion layer combine the loudspeaker scores, which are then
added to the coupling scores. Separately, a small head tries to find the correct
candidate from the audio alone, behind a gradient-reversal layer
\citep{ganin2015}. Any audio cue it can read is thereby trained out of the audio
encoder. This matters because the probe of Section~\ref{sec:failures}(i) is
linear and can miss cues that a network finds.

\paragraph{Training objective.}
Let a batch hold $B$ windows from one listener. For window $i$, $\ell_i\in
\mathbb{R}^K$ are the final candidate scores, $y_i$ is the correct candidate,
$r_i$ is the attended loudspeaker, $e_i$ is the EEG embedding and
$a_i^{+}=\hat a_{i,y_i}$ is the embedding of the correct candidate. We write
$s(e,a)=\tau\,w^{\!\top}c(e,a)$ for the coupling score of Eq.~\ref{eq:coupling}
before centring over candidates, and $\rho$ for a shift of the batch by one
window. \merit{} minimises
\begin{equation}
  \mathcal{L}=\mathcal{L}_{\text{task}}
  +0.3\,\mathcal{L}_{\text{aux}}+\mathcal{L}_{\text{con}}
  +0.5\,\mathcal{L}_{\text{null}}+0.1\,\mathcal{L}_{\text{col}}
  +0.3\,\mathcal{L}_{\text{adv}},
  \label{eq:loss}
\end{equation}
with the six terms
\begin{align*}
  \mathcal{L}_{\text{task}} &= \tfrac1B\textstyle\sum_i
      \mathrm{CE}_{0.1}\big(\ell_i,\,y_i\big),
  \qquad
  \mathcal{L}_{\text{aux}} = \tfrac{1}{B|\mathcal{O}|}\textstyle\sum_i\sum_{m\in\mathcal{O}}
      \mathrm{CE}\big(u_i^m,\,r_i\big),\\
  \mathcal{L}_{\text{con}} &= -\tfrac{1}{2B}\textstyle\sum_i\Big[
      \log\frac{e^{S_{ii}}}{\sum_{j}e^{S_{ij}}}
      +\log\frac{e^{S_{ii}}}{\sum_{j}e^{S_{ji}}}\Big],
  \qquad S_{ij}=s\big(e_i,a_j^{+}\big),\\
  \mathcal{L}_{\text{null}} &= \tfrac1B\textstyle\sum_i\Big[
      \mathrm{sp}\big(s(e_{\rho(i)},a_i^{+})-s(e_i,a_i^{+})+\mu\big)
      +\mathrm{sp}\big(s(\mathbf{0},a_i^{+})-s(e_i,a_i^{+})+\mu\big)\Big],\\
  \mathcal{L}_{\text{col}} &= \tfrac{1}{BD}\textstyle\sum_{i,d}
      \max\big(0,\;\gamma-\mathrm{std}_t\,e_{i,t,d}\big)
      +\tfrac{0.04}{D}\textstyle\sum_{d\neq d'}C_{dd'}^2,\\
  \mathcal{L}_{\text{adv}} &= \tfrac1B\textstyle\sum_i
      \mathrm{CE}\big(\big[g(R(f_{i,k}))\big]_{k=1}^{K},\,y_i\big),
  \qquad f_{i,k}=\big[\,\mathrm{mean}_t\,\hat a_{i,k}\,;\,\mathrm{std}_t\,\hat a_{i,k}\big].
\end{align*}
Here $\mathrm{CE}_{0.1}$ is cross-entropy with label smoothing $0.1$ and
$\mathrm{sp}$ is softplus. $\mathcal{L}_{\text{task}}$ is the decision loss:
the cross-entropy between the $K$ final candidate scores $\ell_i$ and the index
$y_i$ of the attended candidate, so it trains everything that feeds those
scores.
$\mathcal{L}_{\text{aux}}$ gives each orientation head $m\in\mathcal{O}$ its own
loudspeaker prediction $u_i^m$, so the strongest stream cannot take all the
gradient. $\mathcal{L}_{\text{con}}$ is a two-way contrastive loss
\citep{oord2018,radford2021clip} that asks which window's EEG goes with which
window's speech; the batch comes from one listener, so the EEG cannot be matched
by listener identity, and a constant encoder cannot lower this loss.
$\mathcal{L}_{\text{null}}$ requires the real EEG to give the correct candidate a
higher coupling score than another window's EEG or a zeroed EEG does, by a margin
$\mu=0.5$. $\mathcal{L}_{\text{col}}$ keeps the spread over time of each embedding
dimension above $\gamma=0.5$ --- the quantity that falls to zero when an encoder
stops using its input --- and discourages dimensions from copying each other,
where $C$ is the covariance of the time-centred embeddings over all windows and
time steps \citep{bardes2022vicreg}. $\mathcal{L}_{\text{adv}}$ trains a small
network $g$ to find the correct candidate from summary statistics of the audio
embeddings alone; $R$ is a gradient-reversal layer \citep{ganin2015}, which
leaves the forward pass unchanged and flips the sign of the gradient that reaches
the audio encoder, so any audio cue $g$ can read is trained out of it.

\paragraph{Tested but not used: a contribution hinge.}
\label{sec:hinges}
The EEG hinge above acts on the coupling score only. A natural next step is to
apply the same test to the \emph{final} decision, for every stream. For each
stream $m$ we tried
\begin{equation}
  \mathcal{L}^{\text{contrib}}_m=
  \mathrm{softplus}\big(\sigma_y(z_{m\leftarrow\rho})-\sigma_y(z)+\delta\big)
  +\mathrm{softplus}\big(\sigma_y(z_{m\leftarrow 0})-\sigma_y(z)+\delta\big),
  \label{eq:hinge}
\end{equation}
where $\sigma_y(\ell)=\ell_y-\frac1K\sum_k\ell_k$ is the score of the correct
candidate minus the mean score, $z_{m\leftarrow\rho}$ replaces stream $m$ with
another window's copy, $z_{m\leftarrow 0}$ replaces it with zeros, and
$\delta\!=\!0.3$ is the margin. The hinge arms add $0.5\sum_m\lambda_m\mathcal{L}^{\text{contrib}}_m$
to Eq.~\ref{eq:loss}. After each epoch the weight of stream $m$ is set from its
validation contribution $c_m$ in the fused model and the contribution
$c_m^{\text{solo}}$ it reaches through its own head alone:
\begin{equation}
  \lambda_m=\mathrm{clip}\Big(\frac{\max(0,\;c_m^{\text{solo}}-c_m)}{c_m^{\text{solo}}},\;0,\;1\Big),
  \label{eq:gate}
\end{equation}
so a stream that can do well alone but is ignored by the fused model is pushed
hardest. The hinge does not raise the brain's share of credit
(Section~\ref{sec:res-hinge}), so \merit{} trains without it; we report it
because the reason it fails is informative.

\paragraph{Choosing checkpoints by contribution.}
\label{sec:selection}
If a shortcut exists, picking the epoch with the best validation accuracy picks
the epoch that uses the shortcut best. We keep instead the epoch that maximises
\begin{equation}
  \big(v(M)-v(\emptyset)\big) + \beta\,\min_m \phi_m,\qquad \beta=1,
  \label{eq:criterion}
\end{equation}
the validation contribution plus the credit of the \emph{weakest} stream. An epoch that is accurate because it rides one stream loses
to a slightly less accurate epoch that uses all of them. We compute both criteria
at every epoch and report both.

\paragraph{Credit: a split that adds up exactly.}
\label{sec:shapley}
The contributions of single streams do not add up to the total, because streams
overlap --- gaze and the foveal embedding both describe where the listener
looks --- and interact. The Shapley value \citep{shapley1953} over subsets of
\emph{intact} streams is the only split that does add up:
\begin{equation}
  \phi_m=\!\!\sum_{T\subseteq M\setminus\{m\}}\!\!
  \frac{|T|!\,(|M|-|T|-1)!}{|M|!}\Big[v(T\cup\{m\})-v(T)\Big],
  \qquad \sum_{m}\phi_m=v(M)-v(\emptyset).
  \label{eq:shapley}
\end{equation}
This is permutation feature importance \citep{breiman2001,fisher2019,covert2020}
applied to whole streams. With $M\!=\!5$ streams there are only $32$ subsets, so
we compute it exactly, and no subset needs the inputs to be encoded again. The
same random shuffles are used for every subset, so differences between subsets
are not shuffle noise.

%% file: sections/05_setup.tex
\section{Experimental setup}
\label{sec:setup}

\paragraph{Data and task.}
MAESTRO \citep{maestro2026} has sixteen listeners and one hundred trials each.
In every trial four talkers speak at once from loudspeakers at fixed positions,
and the listener is told beforehand which one to attend to. Trials last about
$30$\,s. Each talker recording is used only once, so no audio is shared between
training and test. The attended talker is about $15$\,dB louder than the others
in every trial. We use five streams, all resampled to $64$\,Hz and standardised
per channel: EEG ($32$ channels), gaze ($6$), head motion ($6$), optical flow
from the scene camera ($4$), and one new stream, a \emph{frozen} V-JEPA\,2 ViT-L
embedding \citep{assran2025vjepa2} of the part of the scene the listener is
looking at (one $1024$-d vector per window, never trained). Signals are cut into
windows of $W$ seconds with hop $H$ ($W\!=\!10$, $H\!=\!5$ unless stated). Each
example holds the streams, $K$ candidate speech envelopes of which one is
correct, the index of the correct one, the loudspeaker each candidate came from,
and the listener's identity. The candidates are placed in random order in every
window, so their position says nothing. At test time the order is fixed by the
window's index, so every model we compare sees exactly the same inputs
(Appendix~\ref{app:data}).

Every result is defined by four choices: the candidate set, the split, the
decision window and the set of streams. Every accuracy comes with its own
shuffled baseline; an accuracy without one is not a result. Besides accuracy,
the baseline and the contribution, we report a permutation $p$-value
$(\#\{v_{\text{null}}\!\ge\!v\}+1)/(n+1)$, whose smallest possible value with
twenty shuffles is $0.048$; the Shapley credit $\phi_m$; the \emph{flip rate},
the fraction of windows whose chosen candidate changes when the stream is
shuffled; accuracy with the stream replaced by zeros; and the \emph{collapse},
the mean cosine similarity between the time-centred embeddings of different
windows, where a value near one means every window produces the same pattern.

Training uses AdamW \citep{loshchilov2019adamw} with learning rate $10^{-3}$,
batches of $32$ windows from one listener, and at most $50$ epochs. Checkpoints
are chosen on validation data by the rule of Section~\ref{sec:selection}, with
three shuffles, and the chosen checkpoint is scored once on the test set with
twenty. \textbf{We did no hyper-parameter search}: every setting was fixed once
in advance (Appendix~\ref{app:hp}), so these numbers are not tuned. To compare
architectures, the first row of our ablation runs the benchmark's model
\citep{maestro2026} --- its encoder, its cosine-then-linear score and plain
cross-entropy --- inside our training loop and splits, so that differences
between rows come from the model and not from the protocol. We run it on both
candidate sets. Neither run is meant to reproduce the benchmark's published
numbers, which use its own training loop, a longer schedule and all streams;
when we cite those numbers we take them from the paper.

\paragraph{Why this dataset.}
Crowding out between streams can only be studied when the streams are recorded
\emph{and} the listener is free to produce them; Section~\ref{sec:related}
reviews why the other public corpora rule this out by design. As far as we know,
MAESTRO is the only public AAD dataset that records eye tracking, head motion
and scene video together with EEG from a listener who is free to move in a room
with several loudspeakers (Appendix~\ref{app:corpora}). The parts of \merit{}
that do not need these streams --- the audio probe and the coupling score ---
work on any matching dataset, and we test them on two
(Section~\ref{sec:res-mm}).

%% file: sections/06_results.tex
\section{Results}
\label{sec:results}

\subsection{What the matching task allows}
\label{sec:res-ladder}

We change one thing at a time, starting from the benchmark's model run in our
training loop. The benchmark itself uses matched candidates and reports positive
contributions. Here we first pair its model with \cUnm{} candidates, the usual
choice in the wider literature. This shows what the task allows; it is not a
reproduction of the benchmark's result.

With \cUnm{} candidates the benchmark's model reaches
\R{ladder_raw/L0_published/within/acc} four-way accuracy, but its contribution is
\R{ladder_raw/L0_published/within/contrib}, its decisions never change when the EEG
is shuffled (flip rate \R{ladder_raw/L0_published/within/flip}), and its
embeddings are nearly identical across windows (collapse
\R{ladder_raw/L0_published/within/collapse}). The accuracy is real, but none of it
comes from the EEG: this is the shortcut of Section~\ref{sec:failures}(ii). A
better encoder raises accuracy to \R{ladder_raw/L1_encoder/within/acc} and leaves
the contribution at \R{ladder_raw/L1_encoder/within/contrib} --- it simply reads
the audio better. A change that raises accuracy is not, by itself, evidence; only
the shuffle test shows this. The time-centred score gives the first non-zero
contribution. With matched candidates the same benchmark model scores
\R{ladder_qm/L0_published/within/acc}, which is chance.

\subsection{More streams, less credit for the brain}
\label{sec:res-hinge}

\begin{table}[t]
\caption{\textbf{More streams give more accuracy, and less credit to the brain.}
Within-listener, \cMat{}, chance $0.25$. The four fused models share
streams, data, folds, architecture and listener adaptation; they differ only in
the contribution hinge and in modality dropout. $\phi_m$ is each stream's Shapley
credit and the five add up to the contribution. The EEG-only model shows how much
credit the brain earns on its own. Appendix~\ref{app:hinge2loso} gives the same
four models on unseen listeners.}
\label{tab:hinge}
\centering\small
\setlength{\tabcolsep}{4pt}
\begin{tabular}{@{}lrrrrrrrr@{}}
\toprule
& & & & \multicolumn{5}{c}{Shapley credit $\phi_m$}\\
\cmidrule(l){5-9}
Model & Accuracy & Null & Contrib. & EEG & gaze & head & flow & fovea \\
\midrule
EEG alone
 & \R{v2onset/V4_subjadapt/within/acc} & \R{v2onset/V4_subjadapt/within/null} & \R{v2onset/V4_subjadapt/within/contrib}
 & \R{v2onset/V4_subjadapt/within/contrib} & --- & --- & --- & --- \\
\midrule
\textbf{\merit{}} (no hinge, no dropout)
 & \textbf{\R{v2hinge2/H4_nohinge_nodrop/within/acc}} & \R{v2hinge2/H4_nohinge_nodrop/within/null} & \textbf{\R{v2hinge2/H4_nohinge_nodrop/within/contrib}} & \R{v2hinge2/H4_nohinge_nodrop/within/shap/eeg} & \R{v2hinge2/H4_nohinge_nodrop/within/shap/gaze} & \R{v2hinge2/H4_nohinge_nodrop/within/shap/imu} & \R{v2hinge2/H4_nohinge_nodrop/within/shap/video} & \R{v2hinge2/H4_nohinge_nodrop/within/shap/fovea} \\
No hinge, with dropout
 & \R{v2hinge2/H2_nohinge_drop/within/acc} & \R{v2hinge2/H2_nohinge_drop/within/null} & \R{v2hinge2/H2_nohinge_drop/within/contrib} & \R{v2hinge2/H2_nohinge_drop/within/shap/eeg} & \R{v2hinge2/H2_nohinge_drop/within/shap/gaze} & \R{v2hinge2/H2_nohinge_drop/within/shap/imu} & \R{v2hinge2/H2_nohinge_drop/within/shap/video} & \R{v2hinge2/H2_nohinge_drop/within/shap/fovea} \\
With hinge, no dropout
 & \R{v2hinge2/H3_hinge_nodrop/within/acc} & \R{v2hinge2/H3_hinge_nodrop/within/null} & \R{v2hinge2/H3_hinge_nodrop/within/contrib} & \R{v2hinge2/H3_hinge_nodrop/within/shap/eeg} & \R{v2hinge2/H3_hinge_nodrop/within/shap/gaze} & \R{v2hinge2/H3_hinge_nodrop/within/shap/imu} & \R{v2hinge2/H3_hinge_nodrop/within/shap/video} & \R{v2hinge2/H3_hinge_nodrop/within/shap/fovea} \\
With hinge, with dropout
 & \R{v2hinge2/H1_hinge_drop/within/acc} & \R{v2hinge2/H1_hinge_drop/within/null} & \R{v2hinge2/H1_hinge_drop/within/contrib} & \R{v2hinge2/H1_hinge_drop/within/shap/eeg} & \R{v2hinge2/H1_hinge_drop/within/shap/gaze} & \R{v2hinge2/H1_hinge_drop/within/shap/imu} & \R{v2hinge2/H1_hinge_drop/within/shap/video} & \R{v2hinge2/H1_hinge_drop/within/shap/fovea} \\
\bottomrule
\end{tabular}
\end{table}

\paragraph{The brain's share falls by almost half.}
Adding the behavioural streams raises accuracy from \R{v2onset/V4_subjadapt/within/acc} to
\R{v2hinge2/H4_nohinge_nodrop/within/acc}. But the EEG's credit falls from \R{v2onset/V4_subjadapt/within/contrib}, when it
is the only stream, to \R{v2hinge2/H4_nohinge_nodrop/within/shap/eeg} in the fused model. More streams have
not taught the model more about the brain; they have let it rely on the brain
less. A joint shuffle cannot see this, because the joint contribution
\emph{rises}, from \R{v2onset/V4_subjadapt/within/contrib} to \R{v2hinge2/H4_nohinge_nodrop/within/contrib}. This is why the
per-stream split is the measurement, and the total is not. In the fused model the
foveal embedding earns \R{v2hinge2/H4_nohinge_nodrop/within/shap/fovea}, more than the EEG: much of the
accuracy is a statement about where the listener was looking.

\paragraph{Putting the test in the loss does not fix it.}
The contribution hinge never raised the brain's credit above the matching model
without it. With modality dropout it lowered the EEG's credit from
\R{v2hinge2/H2_nohinge_drop/within/shap/eeg} to \R{v2hinge2/H1_hinge_drop/within/shap/eeg}; without dropout it made no real
difference (\R{v2hinge2/H3_hinge_nodrop/within/shap/eeg} against \R{v2hinge2/H4_nohinge_nodrop/within/shap/eeg}). The reason is the
interaction. Dropout removes streams at random during training, so the
validation contribution that sets the hinge's weights is measured on a model that
keeps being asked to work without them, and the hinge then pushes whichever
stream looks unused --- rarely the EEG. On unseen listeners the models without the
hinge keep more credit on the brain in both pairs (Appendix~\ref{app:hinge2loso}).
A measurement that cannot be pushed up by training against it is one worth
trusting. The one training choice that did raise the brain's credit was choosing
checkpoints by contribution: in an earlier version of this experiment, at
$W=10$\,s, it moved the
EEG's credit from \R{hinge/all_nohinge_accsel/within/shap/eeg} to
\R{hinge/all_nohinge_credsel/within/shap/eeg} (Table~\ref{tab:hingew10}).

\paragraph{Which model we call \merit{}.}
\merit{} is the model without the hinge and without dropout. It is not the most
accurate of the four in either protocol. It keeps the most credit on the brain on
unseen listeners (\R{v2hinge2loso/H4_nohinge_nodrop/loso/shap/eeg}) and ties for the most within listeners
--- and credit is what we select for. Choosing it costs less than $0.02$ in
accuracy in each protocol.

\subsection{The full system, and the benchmark}
\label{sec:res-full}

The results above make one decision per window. But the listener attends to one
talker for the whole trial, so Table~\ref{tab:full} adds, one at a time: one
decision per trial, made by summing the window scores; training on random crops
of the whole trial; limiting the coupling to delays after the sound; a small
per-listener correction of the EEG channels; and a second audio channel that marks
speech onsets. The candidates stay matched and the splits stay the same.

\begin{table}[t]
\caption{\textbf{Improvement ladder}, EEG only, within-listener, \cMat{},
chance $0.25$. Each row adds one change to the row above. The last row is a
control, not a step: it limits the coupling to delays \emph{before} the sound,
where no brain response to that sound can be.}
\label{tab:full}
\centering\small
\begin{tabular}{@{}lrrr@{}}
\toprule
& Accuracy & Null & Contribution \\
\midrule
Random crops, one decision per window & \R{v2onset/V1_crops/within/acc} & \R{v2onset/V1_crops/within/null} & \R{v2onset/V1_crops/within/contrib} \\
\;+ one decision per trial         & \R{v2onset/V2_trial/within/acc} & \R{v2onset/V2_trial/within/null} & \R{v2onset/V2_trial/within/contrib} \\
\;+ only delays after the sound    & \R{v2onset/V3_causal/within/acc} & \R{v2onset/V3_causal/within/null} & \R{v2onset/V3_causal/within/contrib} \\
\textbf{\merit{}, EEG only} (+ per-listener correction) & \textbf{\R{v2onset/V4_subjadapt/within/acc}} & \R{v2onset/V4_subjadapt/within/null} & \textbf{\R{v2onset/V4_subjadapt/within/contrib}} \\
\midrule
\;\emph{control:} only delays before the sound & \R{v2onset/V3b_acausal/within/acc} & \R{v2onset/V3b_acausal/within/null} & \R{v2onset/V3b_acausal/within/contrib} \\
\bottomrule
\end{tabular}
\end{table}

\paragraph{What each change is worth.}
One decision per trial is by far the largest gain: the contribution rises from
\R{v2onset/V1_crops/within/contrib} to \R{v2onset/V2_trial/within/contrib} on the
same windows. Limiting the coupling to delays after the sound adds a little, and
the control confirms the direction: delays before the sound reach only
\R{v2onset/V3b_acausal/within/contrib}. The per-listener correction raises the
contribution from \R{v2onset/V3_causal/within/contrib} to
\R{v2onset/V4_subjadapt/within/contrib}, and the onset channel raises accuracy
from \R{v2ladder/V4_subjadapt/within/acc} to \R{v2onset/V4_subjadapt/within/acc}.

\paragraph{Against the benchmark.}
We compare with the benchmark at its longest window, $30$\,s, which --- like our
per-trial decision --- gives one decision per trial \citep{maestro2026}. There it
reports $0.591$ for EEG alone within listeners and $0.619$ on unseen listeners,
and $0.698$ and $0.674$ for its best multimodal models; its contributions are
$+0.320$ and $+0.363$ for EEG alone, and $+0.413$ for all streams on unseen
listeners. \merit{} reaches \R{v2onset/V4_subjadapt/within/acc} for EEG alone within listeners
(contribution \R{v2onset/V4_subjadapt/within/contrib}), \R{v2hinge2/H4_nohinge_nodrop/within/acc} with all streams
(\R{v2hinge2/H4_nohinge_nodrop/within/contrib}), and \R{v2hinge2loso/H4_nohinge_nodrop/loso/acc} with all streams on unseen
listeners (\R{v2hinge2loso/H4_nohinge_nodrop/loso/contrib}, positive in \R{v2hinge2loso/H4_nohinge_nodrop/loso/pos} folds). Three of
the four comparisons improve on both accuracy and contribution.

The fourth, EEG alone on unseen listeners, does not: \R{v2bestloso/eeg/loso/acc}
against $0.619$, with contribution \R{v2bestloso/eeg/loso/contrib}. The per-listener
correction behind our largest within-listener gain cannot, by design, help a
listener it has never seen. Our gains on the \emph{brain} signal are therefore
shown within listeners.

\subsection{Is the EEG signal real?}
\label{sec:res-mm}

Four checks, with details in Appendix~\ref{app:more}.
\textbf{Same-talker candidates.} When all candidates come from the attended
talker's own speech, the audio alone is at chance and no orienting cue can help;
EEG alone still contributes \R{mm3/eeg/within/contrib} at $K\!=\!3$ and
\R{mm2/eeg/within/contrib} at $K\!=\!2$.
\textbf{Direction in time.} Limited to delays of $+125$ to $+1109$\,ms after the
sound, where a brain response must be, the contribution is
\R{lag_qm/causal/within/contrib}; the mirror image before the sound gives
\R{lag_qm/acausal/within/contrib}. This rules out an electrical echo of the sound
at zero delay, and slow drift that is the same in both directions.
\textbf{Window length.} The contribution grows with the window, as it should for
this task \citep{geirnaert2021} (Figure~\ref{fig:window}). At short windows the
behavioural streams take over: at $2$\,s the EEG alone contributes
\R{w2/eeg/within/contrib} while the foveal embedding alone contributes
\R{w2/fovea/within/contrib} (Appendix~\ref{app:shortwin}).
\textbf{Other datasets.} On two public two-talker datasets the audio probe is at
chance on DTU but \R{extprobe/kul/w10/excess} above chance on KU~Leuven,
because one of its two talkers is the attended one in \R{extprior/kul/nrole} of \R{extprior/kul/ntrials} trials
(Appendix~\ref{app:external}). On KU~Leuven and DTU the coupling branch reaches
\R{extkulwithinst/eeg_centred/within/acc} and \R{extdtuwithinst/eeg_centred/within/acc}
even with same-talker candidates, but the direction-in-time check does not pass:
delays before and after the sound score the same. We therefore do not count these
as brain decoding (Appendix~\ref{app:externalcoupling}). At the $5$\,s window used by recent work,
the coupling branch is more accurate on unseen listeners of both datasets than
NeuroToken \citep{alavi2026neurotoken}, the strongest prior result without data
leakage (Table~\ref{tab:published}). A plain accuracy report
would have done so.

%% file: sections/08_limitations.tex
\section{Conclusion}
\label{sec:conclusion}

A decoder can become more accurate because it has stopped using the input the
claim is about. We showed three ways this happens and built a decoder in which
two of them are impossible. For the third, we made each stream's credit a
number that is measured, used to choose models, and reported for every result.
It shows that adding behavioural streams almost halves the brain's share of the
credit, and that on other public datasets, high accuracy can hide an imbalance
in the labels or a signal that cannot be confirmed as neural. Training against
the measurement did not raise the brain's share. That the measurement resists
this is the best evidence we have that it is measuring something real.

%% file: sections/09_statements.tex
\subsection*{AI use statement}

Generative AI tools were used substantively in this work and their use is
disclosed here in full. An AI coding assistant was used to write the decoder,
data and evaluation code in \texttt{src/} and to produce the first draft of this
manuscript's text, working from the authors' prior analyses, design decisions and
experimental results. It was not used to generate research ideas, to select which
hypotheses to test, or to produce any experimental result: every number reported
here comes from a training run whose configuration, execution and outputs are
recorded on disk, and the manuscript's numbers are inserted mechanically from
those outputs rather than written by hand, so that an unsupported cell is visible
rather than silent (Appendix~\ref{app:repro}). The authors have read and verified
the code, re-derived the coupling-score and Shapley claims of
Section~\ref{sec:method} independently, and checked the cited literature against the
primary sources. The authors take full responsibility for the final content of
this work, including all text, claims and artifacts.

\subsection*{Ethics statement}

The dataset was collected under institutional review with informed consent, and
we use it as released; no new data from people was collected for this work. Some
of the present authors released the benchmark we build on. This gave us access to
its code and to its authors' own account of its limits, and readers should keep
it in mind when reading our comparisons. When we cite the benchmark's numbers we
take them from its paper, rather than re-running it in a setting we chose.
Attention decoding is meant for hearing devices. We state clearly that the
behavioural streams measure where a listener faces, not what they attend to
(Section~\ref{sec:intro}), because a device that confused the two would fail
exactly for the listeners who cannot turn their head or eyes.

\subsection*{Reproducibility statement}

The decoder, the data code and all tests form one package with one configuration
file, and every experiment in the paper is one command (Appendix~\ref{app:repro}).
All settings are listed in Appendix~\ref{app:hp}, and none were tuned. At test
time the order of the candidates is fixed by each window's index, so every model
we compare sees exactly the same inputs, and the models inside one run share one
dataset and one set of folds. Each test set is scored once, and the shuffled
baseline behind every reported number is never used for training or for choosing
a checkpoint.

%% file: sections/A_failures.tex
\section{More detail on the three problems}
\label{app:failures}

This section gives the steps behind Section~\ref{sec:failures} that the main text
states briefly.

\subsection{Why making the candidates equally loud is not enough}
\label{app:conf}

Every step of the envelope pipeline is linear, so scaling a recording by $\alpha$
scales its envelope by $\alpha$: $\mathrm{env}(\alpha x)=\alpha\,\mathrm{env}(x)$.
Standardising each candidate then subtracts its mean and divides by its standard
deviation, which removes any change of the form $x\mapsto\alpha x+\beta$. Playing a
recording $15$\,dB louder or quieter therefore gives exactly the same
standardised envelope. A pure difference in loudness cannot survive, so it cannot
be what gives the answer away.

What does survive is every property that scaling and shifting leave unchanged.
Standardising fixes only two numbers per candidate, the mean and the variance.
Skew, kurtosis, sparsity, ratios between percentiles and the balance between
frequency bands all pass through untouched. If the attended and ignored
recordings differ in any of these, no rescaling can help --- and on this dataset
they do: the envelope's peak-to-average ratio alone separates attended and ignored
talkers with an effect size of $d=\R{sep/envelopecrestdB/d}$. A fix must therefore either make the candidates'
whole distributions identical (\cMat{}) or remove the difference between
attended and ignored recordings altogether (same-talker candidates). Reporting
chance does not help either, since the mark lifts accuracy well above it.

\subsection{How the standard score reduces to an audio classifier}
\label{app:degenerate}

The standard score has one convolution stack for the recording and one, shared
across candidates, for the audio. At each time step $t$ it normalises the brain
embedding $c_t$ and the candidate embedding $a_{k,t}$ to unit length, multiplies
them element by element, averages over time and applies a linear layer:
$\mathrm{score}_k = w^\top\big(\frac1T\sum_t \hat c_t\odot\hat a_{k,t}\big)+b$.
Now let the brain encoder output the same unit vector at every step,
$\hat c_t=\hat c$. The brain term no longer depends on $t$, so it moves out of the
average:
\begin{equation}
  \mathrm{score}_k \;=\; w^\top\!\Big(\hat c \odot \tfrac1T\textstyle\sum_t \hat a_{k,t}\Big) + b
  \;=\; \tilde w^\top \bar a_k + b,\qquad \tilde w=w\odot\hat c .
\end{equation}
This is a linear classifier on the time-averaged audio embedding $\bar a_k$ of
each candidate, and it can read any audio mark the audio encoder learns to
represent. The model reaches it simply by letting the brain encoder go constant,
which is easier than learning to follow the speech. The time-centred score of
Eq.~\ref{eq:coupling} closes this route: with a constant brain embedding,
$e-\bar e=0$, every correlation is zero, and every candidate gets the same score.

\subsection{Why a joint shuffle cannot see crowding out}
\label{app:freeload}

Shuffling all recorded streams together measures how much the model uses the
listener's signals as a whole. Suppose EEG alone earns a contribution of $0.40$.
Now add a gaze stream that is easier to read, and suppose the fused model earns
$0.55$ in total, but only $0.20$ of it from the EEG. The joint contribution went
\emph{up}, from $0.40$ to $0.55$, while the brain's share fell by half. A joint
shuffle reports the first change and hides the second. Only a split per stream,
such as the Shapley values of Eq.~\ref{eq:shapley}, shows both. The numbers of
Table~\ref{tab:hinge} follow exactly this pattern.

\subsection{Why we shuffle the recording, not the labels}
\label{app:perm}

A common check shuffles the labels and retrains or re-scores. That is the wrong
test here. Shuffling the labels breaks the link between the audio and the answer
as well as the link between the brain and the answer, so a decoder that reads
only the audio also drops to chance, and the shortcut we are testing for is
hidden. We instead shuffle the \emph{recording} across test windows and keep each
window's own candidates and label. A decoder that reads the audio keeps its
accuracy; only a decoder that uses the recording loses any. The accuracy that
remains, $v(\emptyset)$, is what \emph{this} model gets from the audio alone,
measured through its own audio pathway. The linear probe bounds what a simple
reader can take from the audio; $v(\emptyset)$ measures what this decoder
actually takes.

%% file: sections/A_related.tex
\section{Related work in more detail}
\label{app:related}

\paragraph{Auditory attention decoding.}
Brain activity follows the envelope of the speech a listener attends to
\citep{ding2012,lalor2009}. Standard decoders either rebuild the envelope from
EEG and correlate it with each candidate \citep{osullivan2015}, or learn the
match directly \citep{accou2021,monesi2020,accou2023,puffay2023}. A separate line
decodes the \emph{direction} of attention without any audio
\citep{vandecappelle2021}. The work closest to ours is \citet{rotaru2024}, who
showed that direction decoding on several public datasets is driven largely by
eye movements and by regularities in the trial design, not by attention. We bring
the same concern to the matching task and to several recorded streams at once,
and we make the check part of how models are chosen and reported rather than a
test run afterwards.

\paragraph{Shortcuts in datasets.}
Models use whatever solves the task \citep{geirhos2020}, and benchmarks often
contain cues that make the intended input unnecessary
\citep{torralba2011,gururangan2018,goyal2017vqa}. The usual fixes are to rebalance
the data \citep{goyal2017vqa} or to train against a branch that sees only the
shortcut \citep{cadene2019rubi}. \merit{} does both --- it checks the candidate
sets and it trains an audio-only adversary \citep{ganin2015} --- and adds one thing
specific to the matching task: a scoring function in which the shortcut is
impossible rather than discouraged (Section~\ref{sec:coupling}).

\paragraph{Balancing modalities.}
When modalities are trained together, the easiest one dominates, and the fused
network can end up worse than its best single part \citep{wang2020gradblend}.
This has been called modality competition \citep{huang2022competition} and a
greedy learning process \citep{wu2022greedy}, and it is usually treated by
evening out \emph{training} --- adjusting gradients as training runs
\citep{peng2022ogm}, blending losses \citep{wang2020gradblend}, or dropping
modalities at random. \merit{} looks at something different. Those methods balance
how fast each modality learns; a scientific claim instead rests on how much each
modality affects the decision \emph{at test time}, which is what
Eq.~\ref{eq:shapley} reports. The two are not the same: in Table~\ref{tab:hinge},
turning modality dropout on or off changes the EEG's credit very little, while the
EEG's credit is still far below what it earns alone.

\paragraph{Attribution.}
Permutation importance \citep{breiman2001,fisher2019} measures how much a model
depends on an input by breaking that input's link to the rest of the example.
Shapley values \citep{shapley1953,lundberg2017} give the only way to split credit
that adds up exactly, and SAGE \citep{covert2020} uses them for importance measured
by performance over a whole dataset. We apply the same idea to whole streams,
where there are few enough subsets ($2^M$) to compute it exactly;
\citet{hessel2020} make a related point for multimodal models. What is new is not
the measurement but how it is used: to choose checkpoints, and to report the
brain's share alongside every accuracy. We also tried adding it to the training
loss, and report that this does not help (Section~\ref{sec:res-hinge}).

\paragraph{Components.}
The contrastive term is InfoNCE \citep{oord2018} in the two-way arrangement of
CLIP \citep{radford2021clip}, with negatives from the same listener; the
anti-collapse term follows VICReg \citep{bardes2022vicreg}; and the foveal stream
is a frozen V-JEPA\,2 encoder \citep{assran2025vjepa2}.

%% file: sections/A_appendix.tex
\section{Checking the candidate sets}
\label{app:probe}

Table~\ref{tab:probe} gives the audio probe for every candidate set. The probe
is a logistic regression on eight statistics of each standardised envelope that
do not change when the signal is rescaled: kurtosis, skew, Gini sparsity, the
range between the 5th and 95th percentiles, the fraction of near-silent samples,
and three frequency-band ratios. It is trained and tested on different audio
content, and picks the candidate it rates most likely to be attended. A candidate
set that does not give the answer away must score $1/K$.

\begin{table}[h]
\caption{\textbf{Audio-only acceptance probe}: what a linear reader takes from
the candidates alone, with no access to any recording. A shortcut-free
construction must sit at chance.}
\label{tab:probe}
\centering\small
\begin{tabular}{@{}llrrr@{}}
\toprule
Construction & & $K$ & Probe & Excess over chance \\
\midrule
Competing talkers                    & \cUnm{}      & 4 & \R{probe/raw/K4}        & \R{probe/raw/K4/excess} \\
\quad + distribution matching        & \cMat{}   & 4 & \R{probe/qmatch/K4}     & \R{probe/qmatch/K4/excess} \\
Same-talker temporal negatives       & \cSame{}    & 3 & \R{probe/shifted/K3}    & \R{probe/shifted/K3/excess} \\
\quad + distribution matching        & \cSameM{} & 3 & \R{probe/shifted_qm/K3} & \R{probe/shifted_qm/K3/excess} \\
Same-talker temporal negatives       & \cSame{}    & 2 & \R{probe/shifted/K2}    & \R{probe/shifted/K2/excess} \\
\quad + distribution matching        & \cSameM{} & 2 & \R{probe/shifted_qm/K2} & \R{probe/shifted_qm/K2/excess} \\
\bottomrule
\end{tabular}
\end{table}

\section{Encoder design}
\label{app:encoder}

Each stream goes through five dilated temporal convolutions
\citep{yu2016dilated} with dilations $1,2,4,8,16$ and GroupNorm
\citep{wu2018groupnorm} after each layer. The EEG first passes through a
$1{\times}1$ convolution that maps the electrodes to $8$ virtual sensors. Three
choices matter. \textbf{The window matches the brain response.} The encoder sees
$63$ samples ($0.98$\,s), about the time over which the cortical response to
speech unfolds \citep{lalor2009}. A much larger window would make the deepest
layers mostly see padding, which is the same in every window, so their output
would stop depending on the input. \textbf{There is no activation on the last
layer.} A rectified output has no negative values, so the cosine between any two
embeddings is squeezed into $[0,1]$, and the correlations the score depends on
would all sit close to one. \textbf{Padding is symmetric.} The brain responds to a
sound $100$ to $300$\,ms after it, so an encoder that looks only backward in time
cannot see the response. One-sided versions are used on purpose to restrict the
model to delays after, or before, the sound (Section~\ref{sec:res-mm}).

\section{Further limitations}
\label{app:limits}

\paragraph{Part of the four-way contribution can come from the trial.}
The attended talker is the same in every window of a trial, so a decoder that
recognises \emph{which trial} it is looking at already knows part of the answer.
The same-talker results do not have this issue, because all candidates there come
from the same trial and the same voice.

\paragraph{One room, sixteen listeners, no tuning.}
The data come from one room, one language and sixteen listeners, and the frozen
video stream in particular reads \emph{this} room. We did not tune any setting.
The numbers are therefore not tuned in our favour, but we also cannot claim they
are the best this model can reach.

\paragraph{The cause of the audio mark is not tested directly.}
Every recording is used in only one role, so the role and the recording cannot be
separated. The dataset's authors explain the mark by clipping during the loudness
adjustment \citep{maestro2026}, but the released data cannot test this directly.
No result depends on it: the probe measures how readable the mark is, whatever
its cause, and both fixes work either way. A dataset in which each recording is
the attended talker in some trials and an ignored one in others would remove the
mark at the source.

\section{Per-stream credit, drawn}
\label{app:creditfig}

\begin{figure}[h]
\centering
\includegraphics[width=0.72\textwidth]{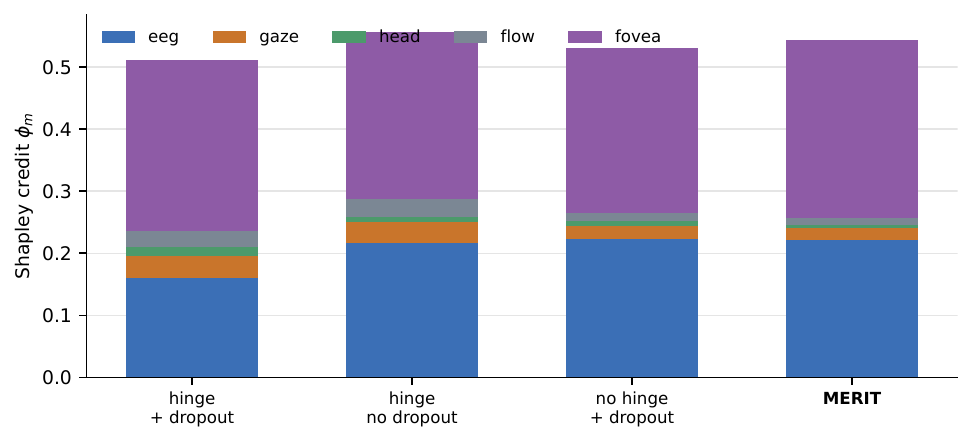}
\caption{\textbf{Shapley credit per stream}, stacked, for the four fused rows
of Table~\ref{tab:hinge}. Each bar's total height is that model's contribution
$v(M)-v(\emptyset)$, decomposed exactly. The question the figure answers is not
how tall the bars are but how much of each one is blue.}
\label{fig:credit}
\end{figure}

\section{Why this dataset, and not another}
\label{app:corpora}

Table~\ref{tab:corpora} compares the public AAD datasets we know of on the two
things that matter here: which streams are recorded next to the EEG, and whether
the listener is free to produce them.

\begin{table}[h]
\caption{\textbf{Public AAD corpora.} ``EOG'' denotes a few electro-oculography
electrodes recorded for artifact rejection, which is not calibrated gaze and
carries no head or scene information. The last column is the one that decides
whether modality freeloading is observable at all.}
\label{tab:corpora}
\centering\small
\resizebox{\textwidth}{!}{%
\begin{tabular}{@{}llllll@{}}
\toprule
Corpus & Gaze & Head & Scene video & Presentation & Listener \\
\midrule
DTU \citep{fuglsang2018dtu}        & 6 EOG & --- & --- & simulated rooms, earphones & soundproof booth \\
Four-speaker \citep{yan2024fourspeaker} & --- & --- & --- & 4 loudspeakers, anechoic & head fixed by chinrest \\
AV-GC-AAD \citep{geirnaert2024avgc} & 4 EOG & --- & --- & earphones, simulated HRTFs & gaze controlled by design \\
\textbf{MAESTRO} \citep{maestro2026} & \textbf{eye tracker} & \textbf{IMU} & \textbf{egocentric} & \textbf{6 loudspeakers, real room} & \textbf{unconstrained} \\
\bottomrule
\end{tabular}}
\end{table}

This pattern is not a flaw in those datasets. It is good design for the question
each one was built to answer. Holding the head and eyes still removes a source of
confusion, and AV-GC-AAD goes further and controls gaze on purpose, so that a
decoder that reads gaze can be caught. That makes it the best available test of
\emph{whether} a decoder uses gaze instead of the brain. But that is a yes-or-no
question about one stream. Our question is about amounts: when a listener
produces five streams at once, how does a fused decoder divide its evidence among
them, and how much stays with the brain? A dataset in which four of the five
streams are held fixed cannot answer it, because the effect has been designed
out. Testing \merit{} there would not be a harder test; it would be a test in
which the effect cannot appear.

Two parts of \merit{} do not need these streams and work on any matching
dataset: the audio probe, which only looks at the candidates, and the coupling
score, whose guarantee is a matter of arithmetic. We test both on two public
datasets in Appendices~\ref{app:external} and~\ref{app:externalcoupling}.

\section{The audio probe on two other datasets}
\label{app:external}

A check that only flags problems in the authors' own dataset is not yet a
reliable tool. We therefore ran the probe of Section~\ref{sec:failures}(i),
unchanged, on two public two-talker datasets collected by other groups:
KU~Leuven \citep{biesmans2017kul,das2019kuldata} and DTU \citep{fuglsang2018dtu},
with $16$ and $18$ listeners. Neither
records gaze, head motion or video, so only the audio part of \merit{} applies ---
which is the point, since the probe depends only on the candidates. We use one
envelope method for all three (power-law compression with exponent $0.6$, a
$20$\,Hz low-pass filter, and resampling to $64$\,Hz), standardise each
candidate, put the candidates in random order, and hold out whole stimulus pairs
for testing. Chance is $0.5$.

\begin{table}[h]
\caption{\textbf{The probe transfers, and it is calibrated.} Audio-only
two-way accuracy on $10$\,s windows, content-disjoint folds, no EEG. The last
column is the corpus's prior on which talker is attended; the probe tracks it.}
\label{tab:external}
\centering\small
\begin{tabular}{@{}lrrrrr@{}}
\toprule
Corpus & Listeners & Windows & Probe & Excess & Attended-role prior \\
\midrule
KU~Leuven & \R{extprobe/kul/w10/nlisteners} & \R{extprobe/kul/w10/nwin} & \textbf{\R{extprobe/kul/w10/probe}} & \textbf{\R{extprobe/kul/w10/excess}} & \R{extprior/kul/prior} \R{extprior/kul/role} \\
DTU & \R{extprobe/dtu/w10/nlisteners} & \R{extprobe/dtu/w10/nwin} & \R{extprobe/dtu/w10/probe} & \R{extprobe/dtu/w10/excess} & \R{extprior/dtu/prior} \R{extprior/dtu/role} \\
\bottomrule
\end{tabular}
\end{table}

\begin{table}[h]
\caption{\textbf{The same probe at four decision windows}, excess over chance.
On DTU, where the attended-role prior is balanced, no cell exceeds $0.005$. The
KU~Leuven excess is present at every window; its $30$\,s cell rests on
\R{extprobe/kul/w30/nwin} windows drawn from eight stimulus pairs and is the least
powered of the four.}
\label{tab:externalsweep}
\centering\small
\begin{tabular}{@{}lrrrr@{}}
\toprule
Corpus & $2$\,s & $5$\,s & $10$\,s & $30$\,s \\
\midrule
KU~Leuven, as released & \R{extprobe/kul/w2/excess} & \R{extprobe/kul/w5/excess} & \textbf{\R{extprobe/kul/w10/excess}} & \R{extprobe/kul/w30/excess} \\
\;\;distribution-matched & \R{extprobe/kul/w2/qmexcess} & \R{extprobe/kul/w5/qmexcess} & \R{extprobe/kul/w10/qmexcess} & \R{extprobe/kul/w30/qmexcess} \\
DTU, as released & \R{extprobe/dtu/w2/excess} & \R{extprobe/dtu/w5/excess} & \R{extprobe/dtu/w10/excess} & \R{extprobe/dtu/w30/excess} \\
\bottomrule
\end{tabular}
\end{table}

\paragraph{One dataset is clean, and that is what gives the other its meaning.}
Across four window lengths (Table~\ref{tab:externalsweep}) the probe stays within
$0.005$ of chance on DTU. The probe is not a warning that goes off everywhere.
Given two talkers who are attended about equally often, a classifier on eight
scale-free statistics cannot tell which one was attended, as it should not be
able to.

\paragraph{On KU~Leuven, the audio alone gives \R{extprobe/kul/w10/excess} above chance.}
On the most widely used AAD benchmark, the same classifier finds the attended
story with accuracy \R{extprobe/kul/w10/probe} on $10$\,s windows, against a
chance of $0.5$, and it is above chance at every window we tried. The cause is
not the audio mark of Section~\ref{sec:failures}(i) but an imbalance in the
labels: track~1 is the attended one in \R{extprior/kul/nrole} of
\R{extprior/kul/ntrials} trials, because the third part of the experiment repeats
the first, and in the first the listener always attends track~1. With two fixed
talkers and one of them attended far more often, the talker's identity
predicts the label, and voice identity is exactly what these statistics carry.
The dataset's own notes warn about two other problems, trial-level overfitting
and gaze; this is a third, separate one. No choice of split can remove it, because
it is part of the design, not of the split.

\paragraph{Matching the distributions only halves it.}
Quantile matching lowers the excess from \R{extprobe/kul/w10/excess} to
\R{extprobe/kul/w10/qmexcess}, and no further. This tells us something. Matching
makes every candidate's set of values the same, which removes the shape
statistics. It keeps each candidate's order in time, so the balance between
frequency bands survives, and with two fixed talkers that balance still reveals
who is speaking. When the problem is a label imbalance between two voices, the
fix is to balance the labels, not to match the distributions. This is a limit of
the candidate set we recommend, found by running our own tool on someone else's
data --- and it is why we report the probe for each dataset instead of calling
one candidate set safe everywhere.

\section{The coupling branch on two other datasets}
\label{app:externalcoupling}

The probe of Appendix~\ref{app:external} needs no model. Here we run the other
part of \merit{} that needs no behavioural stream --- the EEG coupling branch,
its shuffled baseline and the direction-in-time check --- on the same two
datasets. It runs through exactly the same code as every MAESTRO result: the
dataset is handed to the runner through the same data interface, and nothing
after that is changed. Candidates are matched, $K=2$, chance is $0.5$, and
$W=10$\,s. \emph{Same talker} replaces the competing talker with the attended
talker's own speech from another part of the same trial, so neither the talker's
identity nor their position can tell the candidates apart.

\begin{table}[h]
\caption{\textbf{What the full battery says about two public datasets.}
Accuracy and contribution for the unconstrained coupling branch, then
contribution with the branch confined to the causal ($+125\dots+1109$\,ms) and
mirrored acausal ($-1109\dots-125$\,ms) lag bands. \emph{Folds+} is for the
unconstrained branch. The last row is MAESTRO's same-talker $K=2$ result, the
reference for what a directional response looks like.}
\label{tab:externalcoupling}
\centering\small
\resizebox{\textwidth}{!}{%
\begin{tabular}{@{}lrrrrl@{}}
\toprule
& Accuracy & Contribution & Causal & Acausal & Folds+ \\
\midrule
KU~Leuven, within, competing & \R{extkulwithin/eeg_centred/within/acc} & \R{extkulwithin/eeg_centred/within/contrib} & \R{extkulwithin/eeg_causal/within/contrib} & \R{extkulwithin/eeg_acausal/within/contrib} & \R{extkulwithin/eeg_centred/within/pos} \\
KU~Leuven, within, same talker & \R{extkulwithinst/eeg_centred/within/acc} & \R{extkulwithinst/eeg_centred/within/contrib} & \R{extkulwithinst/eeg_causal/within/contrib} & \R{extkulwithinst/eeg_acausal/within/contrib} & \R{extkulwithinst/eeg_centred/within/pos} \\
KU~Leuven, LOSO, competing & \R{extkulloso/eeg_centred/loso/acc} & \R{extkulloso/eeg_centred/loso/contrib} & \R{extkulloso/eeg_causal/loso/contrib} & \R{extkulloso/eeg_acausal/loso/contrib} & \R{extkulloso/eeg_centred/loso/pos} \\
\midrule
DTU, within, competing & \R{extdtuwithin/eeg_centred/within/acc} & \R{extdtuwithin/eeg_centred/within/contrib} & \R{extdtuwithin/eeg_causal/within/contrib} & \R{extdtuwithin/eeg_acausal/within/contrib} & \R{extdtuwithin/eeg_centred/within/pos} \\
DTU, within, same talker & \R{extdtuwithinst/eeg_centred/within/acc} & \R{extdtuwithinst/eeg_centred/within/contrib} & \R{extdtuwithinst/eeg_causal/within/contrib} & \R{extdtuwithinst/eeg_acausal/within/contrib} & \R{extdtuwithinst/eeg_centred/within/pos} \\
DTU, LOSO, competing & \R{extdtuloso/eeg_centred/loso/acc} & \R{extdtuloso/eeg_centred/loso/contrib} & \R{extdtuloso/eeg_causal/loso/contrib} & \R{extdtuloso/eeg_acausal/loso/contrib} & \R{extdtuloso/eeg_centred/loso/pos} \\
\midrule
\midrule
\emph{MAESTRO, within, same talker} & \R{lag_mm/centred/within/acc} & \R{lag_mm/centred/within/contrib} & \textbf{\R{lag_mm/causal/within/contrib}} & \textbf{\R{lag_mm/acausal/within/contrib}} & \R{lag_mm/centred/within/pos} \\
\bottomrule
\end{tabular}}
\end{table}

\paragraph{The coupling branch transfers, and passes the hardest test we have.}
On KU~Leuven and DTU the coupling branch reaches
\R{extkulwithin/eeg_centred/within/acc} and \R{extdtuwithin/eeg_centred/within/acc}
within listeners, positive in every fold, and holds up on unseen listeners
(\R{extkulloso/eeg_centred/loso/acc} and \R{extdtuloso/eeg_centred/loso/acc},
positive in \R{extkulloso/eeg_centred/loso/pos} and
\R{extdtuloso/eeg_centred/loso/pos} folds). The same-talker test is the one that
matters. It removes talker identity, and with it a route that needs no timing at
all: the EEG could reveal \emph{which side} was attended --- through brain
activity that leans to one side, or through small eye movements, which are known
to affect datasets of this kind \citep{rotaru2024} --- and the audio could reveal
which talker is which. Accuracy does not fall under this test; it rises, to
\R{extkulwithinst/eeg_centred/within/acc} and \R{extdtuwithinst/eeg_centred/within/acc}.
Whatever the branch reads, it reads it from the timing.

\paragraph{But the direction-in-time check does not confirm it is the brain.}
On MAESTRO, limiting the coupling to delays after the sound gives about twice the
contribution of the mirror image before the sound
(\R{lag_mm/causal/within/contrib} against \R{lag_mm/acausal/within/contrib}), as a
response to the sound must. On KU~Leuven and DTU the two are the same in every row
of Table~\ref{tab:externalcoupling}. A match that depends on timing but not on its
direction is what a signal at \emph{zero} delay would give, spread equally into
both directions by the slow changes of the speech envelope itself. Both datasets
play the audio through in-ear phones close to the electrodes, and an electrical
leak of the sound into the recording is the classic zero-delay artifact; MAESTRO
plays it through loudspeakers across a room, and there the check passes. We think
this is the most likely explanation, but we have not shown it, and we do not count
the KU~Leuven or DTU accuracies as brain decoding.

\paragraph{Against published results, at the same window.}
Accuracy on these two datasets depends strongly on how the data are split.
\citet{alavi2026neurotoken} re-ran five recent decoders on KU~Leuven at $5$\,s,
once with each paper's own split and once with a split that holds out whole
trials, allows no overlap between training and test windows, and fits all
preprocessing on training trials only. Table~\ref{tab:published} gives the result
next to NeuroToken's own numbers and ours, all at $5$\,s. Chance is $0.5$.

\begin{table}[h]
\caption{\textbf{Two-talker accuracy at $5$\,s: with each paper's own split, and
without leakage.} The first five rows decode the \emph{direction} of attention
from EEG alone, and are NeuroToken's re-runs on KU~Leuven listeners $1$--$4$
\citep{alavi2026neurotoken}; the headline numbers in those papers are higher
still, $95$--$97\%$ at $1$\,s. NeuroToken and \merit{} decode the attended
\emph{talker} from EEG and the candidate speech. NeuroToken holds out whole trials;
\merit{} holds out whole stimuli and uses matched candidates. ``Unseen'' means
listeners not seen in training.}
\label{tab:published}
\centering\small
\begin{tabular}{@{}llrrr@{}}
\toprule
& & \multicolumn{2}{c}{KU~Leuven} & DTU \\
\cmidrule(lr){3-4}\cmidrule(lr){5-5}
Method & Decodes & own split & no leakage & no leakage \\
\midrule
DARNet \citep{yan2024darnet}         & direction & $0.895$ & $0.552$ & --- \\
ListenNet \citep{fan2025listennet}   & direction & $0.930$ & $0.484$ & --- \\
DBPNet \citep{ni2024dbpnet}          & direction & $0.884$ & $0.483$ & --- \\
DenseNet-3D \citep{xu2024densenet}   & direction & $0.864$ & $0.698$ & --- \\
SWIM \citep{zhang2025swim}           & direction & $0.644$ & $0.353$ & --- \\
\midrule
NeuroToken, within listeners         & talker & --- & $0.864$ & $0.837$ \\
NeuroToken, unseen listeners         & talker & --- & $0.634$ & $0.592$ \\
\midrule
\merit{}, within listeners           & talker & --- & \R{extkulwithin5/eeg_centred/within/acc} & \R{extdtuwithin5/eeg_centred/within/acc} \\
\merit{}, unseen listeners           & talker & --- & \R{extkulloso5/eeg_centred/loso/acc} & \R{extdtuloso5/eeg_centred/loso/acc} \\
\bottomrule
\end{tabular}
\end{table}

\paragraph{Reading the table.}
With each paper's own split, the direction decoders reach $0.64$ to $0.93$. Once
leakage is removed they fall by $17$ to $45$ points, to between $0.35$ and $0.70$,
and several land near or below chance: the published numbers mostly measured the
split, not the decoder. NeuroToken is the strongest prior result that survives a
split without leakage. \textbf{On unseen listeners, \merit{} is more accurate than
NeuroToken on both datasets}: \R{extkulloso5/eeg_centred/loso/acc} on
KU~Leuven and \R{extdtuloso5/eeg_centred/loso/acc} on DTU
(Table~\ref{tab:published}), and positive for every one of the
\R{extkulloso5/eeg_centred/loso/nfolds} and \R{extdtuloso5/eeg_centred/loso/nfolds}
held-out listeners. Within listeners it is not: NeuroToken is more accurate than
our \R{extkulwithin5/eeg_centred/within/acc} and
\R{extdtuwithin5/eeg_centred/within/acc} on both, and DenseNet-3D is also higher
on the four KU~Leuven listeners it was run on. The five direction decoders were re-run within
listeners only, so they have no unseen-listener number to compare. Two differences from our setting remain. First, NeuroToken holds
out whole trials, while we hold out whole stimuli. KU~Leuven repeats stories across
trials --- its third part replays the first --- so a split by trial can still put
the same story in training and test, and a split by stimulus cannot. Second, our
candidates are matched, while on KU~Leuven the candidates as released let the audio
alone reach \R{extprobe/kul/w5/excess} above chance at $5$\,s
(Table~\ref{tab:externalsweep}). For our own rows, the caution of this section still
applies: on these two datasets the direction-in-time check does not confirm that
the coupling branch reads a brain response. We have not run that check on the
other methods.

\paragraph{This is the argument of the paper, on other people's data.}
A standard evaluation would report $0.82$ on DTU and stop there. The audio probe
clears the candidates; the same-talker test rules out identity and position; and
the direction check then declines to confirm what is left. Each step is cheap, and
without the last one the number would have been reported as brain decoding.

\section{The hinge experiment on unseen listeners}
\label{app:hinge2loso}

\begin{table}[h]
\caption{\textbf{The $2\times2$ of Table~\ref{tab:hinge}, both protocols.}
$\phi_{\text{EEG}}$ is the EEG's Shapley credit; \emph{Folds+} counts
leave-one-listener-out folds with a positive contribution. No arm is best in
every column.}
\label{tab:hinge2loso}
\centering\small
\begin{tabular}{@{}lrrrrl@{}}
\toprule
& \multicolumn{2}{c}{Within-subject} & \multicolumn{3}{c}{Leave-one-listener-out} \\
\cmidrule(lr){2-3}\cmidrule(lr){4-6}
Arm & Accuracy & $\phi_{\text{EEG}}$ & Accuracy & $\phi_{\text{EEG}}$ & Folds+ \\
\midrule
Hinge, modality dropout & \R{v2hinge2/H1_hinge_drop/within/acc} & \R{v2hinge2/H1_hinge_drop/within/shap/eeg} & \R{v2hinge2loso/H1_hinge_drop/loso/acc} & \R{v2hinge2loso/H1_hinge_drop/loso/shap/eeg} & \R{v2hinge2loso/H1_hinge_drop/loso/pos} \\
No hinge, modality dropout & \R{v2hinge2/H2_nohinge_drop/within/acc} & \R{v2hinge2/H2_nohinge_drop/within/shap/eeg} & \R{v2hinge2loso/H2_nohinge_drop/loso/acc} & \R{v2hinge2loso/H2_nohinge_drop/loso/shap/eeg} & \R{v2hinge2loso/H2_nohinge_drop/loso/pos} \\
\textbf{\merit{}} (no hinge, no dropout) & \R{v2hinge2/H4_nohinge_nodrop/within/acc} & \R{v2hinge2/H4_nohinge_nodrop/within/shap/eeg} & \R{v2hinge2loso/H4_nohinge_nodrop/loso/acc} & \R{v2hinge2loso/H4_nohinge_nodrop/loso/shap/eeg} & \R{v2hinge2loso/H4_nohinge_nodrop/loso/pos} \\
Hinge, no dropout & \R{v2hinge2/H3_hinge_nodrop/within/acc} & \R{v2hinge2/H3_hinge_nodrop/within/shap/eeg} & \R{v2hinge2loso/H3_hinge_nodrop/loso/acc} & \R{v2hinge2loso/H3_hinge_nodrop/loso/shap/eeg} & \R{v2hinge2loso/H3_hinge_nodrop/loso/pos} \\
\bottomrule
\end{tabular}
\end{table}

The order of the four models changes between the two splits. The model with the
hinge and no dropout is the most accurate within listeners and the least accurate
on unseen ones. In both pairs, the model without the hinge keeps more credit on
the brain on unseen listeners. \merit{} is the model without the hinge and
without dropout: it has the most EEG credit on unseen listeners and ties for the
most within listeners.

\section{Every stream set, both splits}
\label{app:loso}

Table~\ref{tab:streams} gives every stream set within listeners, and
Table~\ref{tab:streamsloso} gives the same models on unseen listeners. Both use
per-window decisions at $W=10$\,s.

\begin{table}[t]
\caption{\textbf{Accuracy, null and contribution by stream set}, within-subject
(five content-disjoint folds), \cMat{}, $W=10$\,s, chance $0.25$.
\emph{Folds+} counts folds with a positive contribution. The
leave-one-listener-out counterpart is Table~\ref{tab:streamsloso}.}
\label{tab:streams}
\centering\small
\begin{tabular}{@{}lrrrrl@{}}
\toprule
Streams & Accuracy & Null & Contribution & $p$ & Folds+ \\
\midrule
EEG & \R{streams/eeg/within/acc} & \R{streams/eeg/within/null} & \R{streams/eeg/within/contrib} & \R{streams/eeg/within/p} & \R{streams/eeg/within/pos} \\
EEG + gaze & \R{streams/eeg_gaze/within/acc} & \R{streams/eeg_gaze/within/null} & \R{streams/eeg_gaze/within/contrib} & \R{streams/eeg_gaze/within/p} & \R{streams/eeg_gaze/within/pos} \\
EEG + head & \R{streams/eeg_imu/within/acc} & \R{streams/eeg_imu/within/null} & \R{streams/eeg_imu/within/contrib} & \R{streams/eeg_imu/within/p} & \R{streams/eeg_imu/within/pos} \\
EEG + flow & \R{streams/eeg_flow/within/acc} & \R{streams/eeg_flow/within/null} & \R{streams/eeg_flow/within/contrib} & \R{streams/eeg_flow/within/p} & \R{streams/eeg_flow/within/pos} \\
EEG + fovea & \R{streams/eeg_fovea/within/acc} & \R{streams/eeg_fovea/within/null} & \R{streams/eeg_fovea/within/contrib} & \R{streams/eeg_fovea/within/p} & \R{streams/eeg_fovea/within/pos} \\
Behaviour only & \R{streams/behaviour/within/acc} & \R{streams/behaviour/within/null} & \R{streams/behaviour/within/contrib} & \R{streams/behaviour/within/p} & \R{streams/behaviour/within/pos} \\
\textbf{All five} & \R{streams/all/within/acc} & \R{streams/all/within/null} & \R{streams/all/within/contrib} & \R{streams/all/within/p} & \R{streams/all/within/pos} \\
\bottomrule
\end{tabular}
\end{table}

\begin{table}[h]
\caption{\textbf{Leave-one-listener-out}, sixteen folds, each with a held-out
listener on held-out content. \cMat{}, $W=10$\,s, chance $0.25$.}
\label{tab:streamsloso}
\centering\small
\begin{tabular}{@{}lrrrrl@{}}
\toprule
Streams & Accuracy & Null & Contribution & $p$ & Folds+ \\
\midrule
EEG & \R{loso/eeg/loso/acc} & \R{loso/eeg/loso/null} & \R{loso/eeg/loso/contrib} & \R{loso/eeg/loso/p} & \R{loso/eeg/loso/pos} \\
EEG + gaze & \R{loso/eeg_gaze/loso/acc} & \R{loso/eeg_gaze/loso/null} & \R{loso/eeg_gaze/loso/contrib} & \R{loso/eeg_gaze/loso/p} & \R{loso/eeg_gaze/loso/pos} \\
EEG + head & \R{loso/eeg_imu/loso/acc} & \R{loso/eeg_imu/loso/null} & \R{loso/eeg_imu/loso/contrib} & \R{loso/eeg_imu/loso/p} & \R{loso/eeg_imu/loso/pos} \\
EEG + flow & \R{loso/eeg_flow/loso/acc} & \R{loso/eeg_flow/loso/null} & \R{loso/eeg_flow/loso/contrib} & \R{loso/eeg_flow/loso/p} & \R{loso/eeg_flow/loso/pos} \\
EEG + fovea & \R{loso/eeg_fovea/loso/acc} & \R{loso/eeg_fovea/loso/null} & \R{loso/eeg_fovea/loso/contrib} & \R{loso/eeg_fovea/loso/p} & \R{loso/eeg_fovea/loso/pos} \\
Behaviour only & \R{loso/behaviour/loso/acc} & \R{loso/behaviour/loso/null} & \R{loso/behaviour/loso/contrib} & \R{loso/behaviour/loso/p} & \R{loso/behaviour/loso/pos} \\
\textbf{All five} & \R{loso/all/loso/acc} & \R{loso/all/loso/null} & \R{loso/all/loso/contrib} & \R{loso/all/loso/p} & \R{loso/all/loso/pos} \\
\bottomrule
\end{tabular}
\end{table}

\section{Checks on the EEG signal}
\label{app:more}

\begin{figure}[h]
\centering
\includegraphics[width=0.86\textwidth]{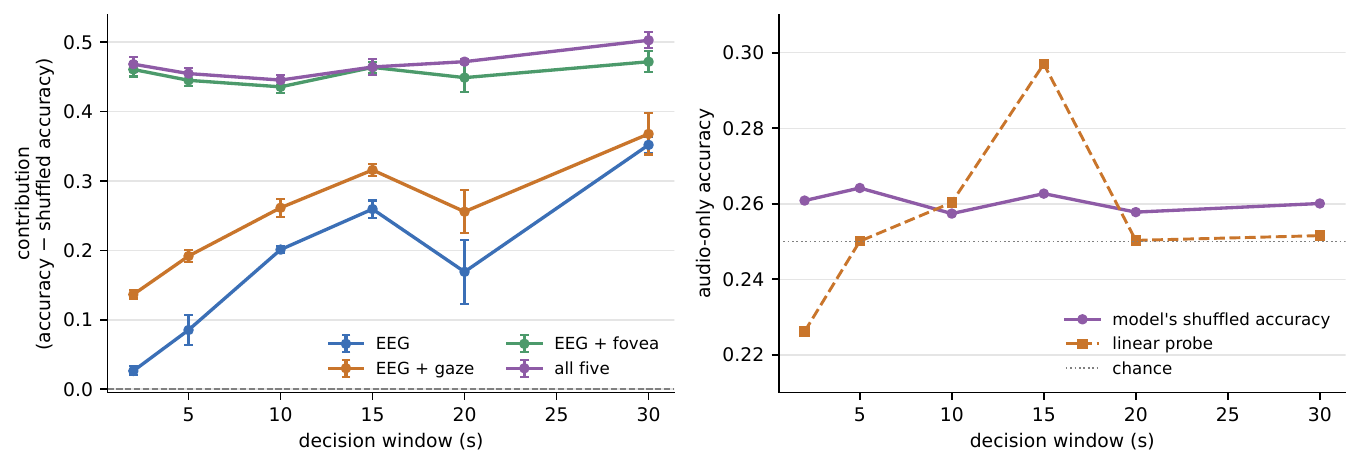}
\caption{\textbf{Contribution against decision-window length}, within-listener,
\cMat{}. Left: contribution for the four stream sets measured at every
window; the eleven stream sets at $2$ and $5$\,s are in Table~\ref{tab:shortwin}.
Right: the full model's shuffled accuracy, which is what the contribution
subtracts, and the linear audio probe, both measured at each window. The model's
shuffled accuracy stays within $0.015$ of chance throughout; the probe varies
more, between $0.23$ and $0.30$.}
\label{fig:window}
\end{figure}

Table~\ref{tab:mm} gives the same-talker results and the direction-in-time
check. Figure~\ref{fig:window} shows how the contribution changes with the
decision window.

\begin{table}[h]
\caption{\textbf{Orienting-free decoding and neural directionality.} EEG only
unless stated, within-subject, $W=10$\,s. Same-talker candidates are the attended
talker at different times, so slot index carries no spatial meaning and no
orienting cue can help. The lag band is set by the encoder's direction plus an
eight-sample shift; a contribution as large in the acausal band would indicate a
shared artifact rather than envelope tracking.}
\label{tab:mm}
\centering\small
\setlength{\tabcolsep}{5pt}
\begin{tabular}{@{}llllrrrr@{}}
\toprule
Candidates & $K$ & Streams & EEG visible at & Chance & Accuracy & Null & Contribution \\
\midrule
\cSameM{} & 3 & EEG      & centred & 0.333 & \R{mm3/eeg/within/acc} & \R{mm3/eeg/within/null} & \textbf{\R{mm3/eeg/within/contrib}}\\
\cSameM{} & 3 & all five & centred & 0.333 & \R{mm3/all/within/acc} & \R{mm3/all/within/null} & \textbf{\R{mm3/all/within/contrib}}\\
\cSameM{} & 2 & EEG      & centred & 0.500 & \R{mm2/eeg/within/acc} & \R{mm2/eeg/within/null} & \textbf{\R{mm2/eeg/within/contrib}}\\
\cSameM{} & 2 & all five & centred & 0.500 & \R{mm2/all/within/acc} & \R{mm2/all/within/null} & \textbf{\R{mm2/all/within/contrib}}\\
\midrule
\cMat{}   & 4 & EEG & $+125\dots+1109$\,ms & 0.250 & \R{lag_qm/causal/within/acc} & \R{lag_qm/causal/within/null} & \textbf{\R{lag_qm/causal/within/contrib}}\\
\cMat{}   & 4 & EEG & $-1109\dots-125$\,ms & 0.250 & \R{lag_qm/acausal/within/acc} & \R{lag_qm/acausal/within/null} & \textbf{\R{lag_qm/acausal/within/contrib}}\\
\cSameM{} & 2 & EEG & $+125\dots+1109$\,ms & 0.500 & \R{lag_mm/causal/within/acc} & \R{lag_mm/causal/within/null} & \textbf{\R{lag_mm/causal/within/contrib}}\\
\cSameM{} & 2 & EEG & $-1109\dots-125$\,ms & 0.500 & \R{lag_mm/acausal/within/acc} & \R{lag_mm/acausal/within/null} & \textbf{\R{lag_mm/acausal/within/contrib}}\\
\bottomrule
\end{tabular}
\end{table}

\subsection{Short decision windows, by stream set}
\label{app:shortwin}

Figure~\ref{fig:window} changes the window for a few stream sets.
Table~\ref{tab:shortwin} does the opposite at the two shortest windows: every
stream set at $W=2$\,s and $W=5$\,s, within listeners, one decision per window,
with half-window hops. The rows extend Figure~\ref{fig:window} downward and can be
compared directly with a benchmark that decides one window at a time. Short
windows are where the problem this paper is about is strongest. The behavioural
streams show \emph{where} the listener faces, which a two-second look already
settles. The coupling branch needs time for a brain response to build up, and its
own window is already $0.98$\,s. The table shows this clearly. At $W=2$\,s the
EEG alone contributes \R{w2/eeg/within/contrib}, while the frozen foveal embedding
alone reaches \R{w2/fovea/within/acc} with a contribution of
\R{w2/fovea/within/contrib}, and the full five-stream model reaches
\R{w2/all/within/acc}. At that window a fused accuracy is almost entirely about
where the listener was looking. It should not be read as a brain result, and
without the per-stream split nothing in the headline number would say so.

\begin{table}[h]
\caption{\textbf{Every stream set at the two shortest decision windows},
within-subject (content-disjoint, five folds), \cMat{}, $K=4$, chance
$0.25$, hop $=W/2$. \emph{Contribution} is accuracy minus the model's own
permutation null; $p$ is the permutation $p$-value, floored at $0.048$ for twenty
permutations. The first block is single streams, the second EEG paired with each
behavioural stream.}
\label{tab:shortwin}
\centering\small
\resizebox{\textwidth}{!}{%
\begin{tabular}{@{}lrrrrrrrr@{}}
\toprule
& \multicolumn{4}{c}{$W=2$\,s} & \multicolumn{4}{c}{$W=5$\,s} \\
\cmidrule(lr){2-5}\cmidrule(lr){6-9}
Streams & Acc. & Null & Contrib. & $p$ & Acc. & Null & Contrib. & $p$ \\
\midrule
EEG & \R{w2/eeg/within/acc} & \R{w2/eeg/within/null} & \R{w2/eeg/within/contrib} & \R{w2/eeg/within/p} & \R{w5/eeg/within/acc} & \R{w5/eeg/within/null} & \R{w5/eeg/within/contrib} & \R{w5/eeg/within/p} \\
gaze & \R{w2/gaze/within/acc} & \R{w2/gaze/within/null} & \R{w2/gaze/within/contrib} & \R{w2/gaze/within/p} & \R{w5/gaze/within/acc} & \R{w5/gaze/within/null} & \R{w5/gaze/within/contrib} & \R{w5/gaze/within/p} \\
head motion & \R{w2/imu/within/acc} & \R{w2/imu/within/null} & \R{w2/imu/within/contrib} & \R{w2/imu/within/p} & \R{w5/imu/within/acc} & \R{w5/imu/within/null} & \R{w5/imu/within/contrib} & \R{w5/imu/within/p} \\
optical flow & \R{w2/flow/within/acc} & \R{w2/flow/within/null} & \R{w2/flow/within/contrib} & \R{w2/flow/within/p} & \R{w5/flow/within/acc} & \R{w5/flow/within/null} & \R{w5/flow/within/contrib} & \R{w5/flow/within/p} \\
fovea (V-JEPA\,2) & \R{w2/fovea/within/acc} & \R{w2/fovea/within/null} & \R{w2/fovea/within/contrib} & \R{w2/fovea/within/p} & \R{w5/fovea/within/acc} & \R{w5/fovea/within/null} & \R{w5/fovea/within/contrib} & \R{w5/fovea/within/p} \\
behaviour (all four) & \R{w2/behaviour/within/acc} & \R{w2/behaviour/within/null} & \R{w2/behaviour/within/contrib} & \R{w2/behaviour/within/p} & \R{w5/behaviour/within/acc} & \R{w5/behaviour/within/null} & \R{w5/behaviour/within/contrib} & \R{w5/behaviour/within/p} \\
\midrule
EEG + gaze & \R{w2/eeg_gaze/within/acc} & \R{w2/eeg_gaze/within/null} & \R{w2/eeg_gaze/within/contrib} & \R{w2/eeg_gaze/within/p} & \R{w5/eeg_gaze/within/acc} & \R{w5/eeg_gaze/within/null} & \R{w5/eeg_gaze/within/contrib} & \R{w5/eeg_gaze/within/p} \\
EEG + head & \R{w2/eeg_imu/within/acc} & \R{w2/eeg_imu/within/null} & \R{w2/eeg_imu/within/contrib} & \R{w2/eeg_imu/within/p} & \R{w5/eeg_imu/within/acc} & \R{w5/eeg_imu/within/null} & \R{w5/eeg_imu/within/contrib} & \R{w5/eeg_imu/within/p} \\
EEG + flow & \R{w2/eeg_flow/within/acc} & \R{w2/eeg_flow/within/null} & \R{w2/eeg_flow/within/contrib} & \R{w2/eeg_flow/within/p} & \R{w5/eeg_flow/within/acc} & \R{w5/eeg_flow/within/null} & \R{w5/eeg_flow/within/contrib} & \R{w5/eeg_flow/within/p} \\
EEG + fovea & \R{w2/eeg_fovea/within/acc} & \R{w2/eeg_fovea/within/null} & \R{w2/eeg_fovea/within/contrib} & \R{w2/eeg_fovea/within/p} & \R{w5/eeg_fovea/within/acc} & \R{w5/eeg_fovea/within/null} & \R{w5/eeg_fovea/within/contrib} & \R{w5/eeg_fovea/within/p} \\
\textbf{all five} & \R{w2/all/within/acc} & \R{w2/all/within/null} & \R{w2/all/within/contrib} & \R{w2/all/within/p} & \R{w5/all/within/acc} & \R{w5/all/within/null} & \R{w5/all/within/contrib} & \R{w5/all/within/p} \\
\bottomrule
\end{tabular}}
\end{table}

\section{The hinge experiment at ten seconds}
\label{app:hingew10}

Table~\ref{tab:hingew10} repeats the hinge experiment at the earlier setting
($W=10$\,s, one decision per window). It also changes how checkpoints are chosen,
which Table~\ref{tab:hinge} does not. Choosing by contribution instead of by
accuracy is the one training choice that raises the EEG's credit.

\begin{table}[h]
\caption{\textbf{Hinge, selection rule and modality dropout at $W=10$\,s}, per-window decisions --- the earlier operating point, kept because it varies the selection rule, which Table~\ref{tab:hinge} does not. Identical
streams, data and folds; within-subject, \cMat{}, $W=10$\,s, chance
$0.25$. $\phi_m$ is Shapley credit and sums to the contribution by construction.
The single-stream EEG model is shown for reference.}
\label{tab:hingew10}
\centering\small
\setlength{\tabcolsep}{4pt}
\begin{tabular}{@{}lrrrrrrrr@{}}
\toprule
& & & & \multicolumn{5}{c}{Shapley credit $\phi_m$}\\
\cmidrule(l){5-9}
Model & Accuracy & Null & Contrib. & EEG & gaze & head & flow & fovea \\
\midrule
EEG alone (reference)
 & \R{streams/eeg/within/acc} & \R{streams/eeg/within/null} & \R{streams/eeg/within/contrib}
 & \R{streams/eeg/within/contrib} & --- & --- & --- & --- \\
\midrule
No hinge, accuracy-selected
 & \R{hinge/all_nohinge_accsel/within/acc} & \R{hinge/all_nohinge_accsel/within/null} & \R{hinge/all_nohinge_accsel/within/contrib}
 & \R{hinge/all_nohinge_accsel/within/shap/eeg} & \R{hinge/all_nohinge_accsel/within/shap/gaze} & \R{hinge/all_nohinge_accsel/within/shap/imu} & \R{hinge/all_nohinge_accsel/within/shap/video} & \R{hinge/all_nohinge_accsel/within/shap/fovea} \\
No hinge, credit-selected
 & \R{hinge/all_nohinge_credsel/within/acc} & \R{hinge/all_nohinge_credsel/within/null} & \R{hinge/all_nohinge_credsel/within/contrib}
 & \R{hinge/all_nohinge_credsel/within/shap/eeg} & \R{hinge/all_nohinge_credsel/within/shap/gaze} & \R{hinge/all_nohinge_credsel/within/shap/imu} & \R{hinge/all_nohinge_credsel/within/shap/video} & \R{hinge/all_nohinge_credsel/within/shap/fovea} \\
Hinge, accuracy-selected
 & \R{hinge/all_hinge_accsel/within/acc} & \R{hinge/all_hinge_accsel/within/null} & \R{hinge/all_hinge_accsel/within/contrib}
 & \R{hinge/all_hinge_accsel/within/shap/eeg} & \R{hinge/all_hinge_accsel/within/shap/gaze} & \R{hinge/all_hinge_accsel/within/shap/imu} & \R{hinge/all_hinge_accsel/within/shap/video} & \R{hinge/all_hinge_accsel/within/shap/fovea} \\
Hinge, credit-selected
 & \R{hinge/all_hinge_credsel/within/acc} & \R{hinge/all_hinge_credsel/within/null} & \R{hinge/all_hinge_credsel/within/contrib}
 & \textbf{\R{hinge/all_hinge_credsel/within/shap/eeg}} & \R{hinge/all_hinge_credsel/within/shap/gaze} & \R{hinge/all_hinge_credsel/within/shap/imu} & \R{hinge/all_hinge_credsel/within/shap/video} & \R{hinge/all_hinge_credsel/within/shap/fovea} \\
\midrule
Hinge, uniform weights
 & \R{hinge/all_hinge_uniform/within/acc} & \R{hinge/all_hinge_uniform/within/null} & \R{hinge/all_hinge_uniform/within/contrib}
 & \R{hinge/all_hinge_uniform/within/shap/eeg} & \R{hinge/all_hinge_uniform/within/shap/gaze} & \R{hinge/all_hinge_uniform/within/shap/imu} & \R{hinge/all_hinge_uniform/within/shap/video} & \R{hinge/all_hinge_uniform/within/shap/fovea} \\
Hinge, no modality dropout
 & \R{hinge/all_hinge_nodrop/within/acc} & \R{hinge/all_hinge_nodrop/within/null} & \R{hinge/all_hinge_nodrop/within/contrib}
 & \R{hinge/all_hinge_nodrop/within/shap/eeg} & \R{hinge/all_hinge_nodrop/within/shap/gaze} & \R{hinge/all_hinge_nodrop/within/shap/imu} & \R{hinge/all_hinge_nodrop/within/shap/video} & \R{hinge/all_hinge_nodrop/within/shap/fovea} \\
No hinge, no modality dropout
 & \R{hinge/all_nohinge_nodrop/within/acc} & \R{hinge/all_nohinge_nodrop/within/null} & \R{hinge/all_nohinge_nodrop/within/contrib}
 & \R{hinge/all_nohinge_nodrop/within/shap/eeg} & \R{hinge/all_nohinge_nodrop/within/shap/gaze} & \R{hinge/all_nohinge_nodrop/within/shap/imu} & \R{hinge/all_nohinge_nodrop/within/shap/video} & \R{hinge/all_nohinge_nodrop/within/shap/fovea} \\
\bottomrule
\end{tabular}
\end{table}

\section{The training objective, explained}
\label{app:objective}

Equation~\ref{eq:loss} gives the loss. This section explains two choices behind
it. \textbf{Why the contrastive term uses one listener at a time.} Raw EEG
identifies the listener with accuracy $0.90$ in a sixteen-way test, against a
chance of $0.0625$. In a batch that mixes listeners, the contrastive term could
match EEG to speech by recognising the listener, which says nothing about
attention. Every batch is therefore drawn from a single listener. \textbf{Why the
contrastive term and the anti-collapse term cannot be satisfied by collapse.} If
the EEG encoder outputs a vector that does not change over time, the time-centred
score gives $s(e,a)=0$ for every pair, so every entry of $S$ in
$\mathcal{L}_{\text{con}}$ is zero and the loss is exactly $\log B$, its value at
chance. And the first part of $\mathcal{L}_{\text{col}}$ measures
exactly the spread over time that such an encoder has lost. \merit{} trains
without modality dropout; the dropout models of Table~\ref{tab:hinge} drop each
stream with probability $0.3$, never all of them at once, and fill a dropped
stream with zeros so the fusion layer always runs.

\section{Reproducibility}
\label{app:repro}

The decoder, the data code and all tests are one package with one configuration
file, and every experiment is one command. The main results come from:

\begin{quote}\small\ttfamily
\# Table 1 and Appendix \ref{app:hinge2loso}: the four fused models, both splits\\
python -m src.main data=merit model=merit runner=merit\_hinge2 \$CROP \textbackslash\\
\quad data.euclidean\_align=true data.audio\_feats=env\_onset\\
python -m src.main data=merit model=merit runner=merit\_hinge2loso0 \$CROP ...\\
\# Table 2: the EEG-only ladder\\
python -m src.main data=merit model=merit runner=merit\_ladder2 \$CROP \textbackslash\\
\quad data.audio\_feats=env\_onset\\
\# Appendices: stream sets, short windows, other datasets\\
python -m src.main data=merit model=merit runner=merit\_streams\\
python -m src.main data=merit model=merit runner=merit\_winmod0 data.window\_sec=2 data.hop\_sec=1\\
python -m src.main data=external model=merit runner=merit\_external data.corpus=dtu\\
python -m src.main mode=certify data=merit model=merit runner=merit
\end{quote}
where \texttt{\$CROP} is \texttt{data.window\_sec=30 data.hop\_sec=15
data.crop\_len=640 data.n\_train\_crops=8 data.n\_eval\_crops=5}. The job scripts
that ran each experiment are released with the code.

The models inside one run share one dataset and one set of folds, so a difference
between two rows comes from the models alone. At test time the order of the
candidates is fixed by each window's index, so every model sees exactly the same
inputs. Every number in the paper is written from the run outputs by a script and
inserted automatically, never typed by hand, and a result is only used once every
fold of its run has finished.

\section{Hyper-parameters}
\label{app:hp}

\begin{table}[h]
\centering\small
\caption{Settings. None were tuned; each was set once from first principles.}
\begin{tabular}{@{}ll@{}}
\toprule
Item & Value \\
\midrule
Sampling rate / window / hop            & $64$\,Hz / $10$\,s / $5$\,s for per-window results (sweep: $2,5,10,15,20,30$\,s at $50\%$ hop) \\
Per-trial decisions (\merit{})          & $10$\,s crops of $30$\,s trials: $8$ random crops per trial when training, $5$ fixed crops at test; window scores summed per trial \\
Stream channels                         & EEG $32$ ($64$ on KU~Leuven and DTU), gaze $6$, head $6$, flow $4$, fovea $1024$ (frozen) \\
Descriptor width $D$ / layers           & $16$ / $5$ (dilations $1,2,4,8,16$; RF $63$ samples $=0.98$\,s)\\
Spatial filters (EEG)                   & $8$\\
Normalisation / activation              & GroupNorm($4$ groups) after every conv; no activation on the last\\
Dropout / modality dropout              & $0.1$ / none in \merit{}; $0.3$ in the dropout models of Table~\ref{tab:hinge} (never all streams; dropped streams filled with zeros) \\
Delays (\merit{})                       & EEG encoder sees only delays after the sound, offset $8$ samples ($+125$ to $+1109$\,ms) \\
Listener correction (\merit{})          & rank-$2$ correction of the EEG channels per listener; in the fused model the EEG is also whitened per listener (not in the EEG-only ladder) \\
Audio channels (\merit{})               & envelope and its rectified derivative (speech onsets) \\
Coupling temperature                    & learned, initialised at $1/0.07$\\
Optimiser / lr / weight decay           & AdamW / $10^{-3}$ / $10^{-4}$; grad-norm clip $1.0$\\
Batch                                   & $32$, drawn from a single listener\\
Schedule                                & lr $\times0.5$ after $5$ epochs without improvement; $\le50$ epochs; patience $12$\\
Label smoothing                         & $0.1$\\
Loss weights (Eq.~\ref{eq:loss})       & task $1.0$ (label smoothing $0.1$); auxiliary $0.3$; contrastive $1.0$; EEG null hinge $0.5$ (margin $\mu=0.5$); anti-collapse $0.1$ ($\gamma=0.5$); adversary $0.3$ \\
Contribution hinge (Eq.~\ref{eq:hinge}) & none in \merit{}; weight $0.5$, margin $\delta=0.3$ in the hinge models \\
Hinge weights                           & Eq.~\ref{eq:gate} in Table~\ref{tab:hinge}; $\lambda_m=\mathrm{clip}((\kappa-c_m)/\kappa,0,1)$, $\kappa=0.05$, in Table~\ref{tab:hingew10} \\
Checkpoint choice                       & Eq.~\ref{eq:criterion}, $\beta=1$ (accuracy recorded as well) \\
Permutations                            & $3$ on inner validation; $20$ at test; $5$ draws per Shapley coalition\\
Splits                                  & within: $5$ content-disjoint folds, $20\%$ inner val (content-disjoint)\\
                                        & LOSO: $16$ folds, $20\%$ global content holdout, inner val listener-disjoint\\
\bottomrule
\end{tabular}
\end{table}

\section{Why the Shapley credits add up}
\label{app:additivity}

Equation~\ref{eq:shapley} guarantees $\sum_m\phi_m=v(M)-v(\emptyset)$ exactly.
The code reports both sides for every run as a check, and they agree to the
printed precision in every result in this paper. All $2^5=32$ subsets are
computed rather than sampled, and the same shuffles are reused across subsets, so
a difference between two subsets is not shuffle noise. The cost is mostly in
re-scoring, not re-encoding, because the stream embeddings and audio embeddings
are computed once per evaluation.

Two cautions. First, $\phi_m$ is the credit \emph{for this trained model on this
test set}, not a fixed property of the stream: a stream may carry information the
model did not use. Second, when two streams overlap --- gaze and the foveal
embedding both describe where the listener looks --- Shapley splits their shared
credit between them. That is the right answer to ``how much would I lose by
removing this one, averaged over the order of removal,'' but it does not mean that
each stream holds half of the information.

\section{Three per-stream measures}
\label{app:perstream}

The runs report three per-stream numbers, which answer different questions. The
\emph{contribution} $v(M)-v(M\setminus\{m\})$ is the accuracy lost by shuffling
stream $m$ while all others stay intact; it is small for a stream whose
information is also in another stream. The \emph{Shapley credit} $\phi_m$ is the
same quantity averaged over all subsets of the other streams, and it is the one
that adds up to the total. \emph{Leave-one-out} is the accuracy lost by
\emph{zeroing} stream $m$; it differs from shuffling when the model uses a
stream's overall level rather than its window-by-window values. All three are in
the released result tables.

\section{Dataset}
\label{app:data}

\paragraph{Design.}
MAESTRO \citep{maestro2026} has sixteen listeners with one hundred trials each.
In every trial four talkers speak at once from loudspeakers at fixed, known
positions. The listener is told before the trial which talker to attend to and
answers a question about it afterwards. Trials last about $30$\,s, and each
loudspeaker is the attended one equally often. The talkers of a trial are
different recordings and no recording is used twice, so no audio is shared
between training and test. The attended talker is played about $15$\,dB louder
than the others in every trial; this is why the difference between loudness and
shape in Section~\ref{sec:failures}(i) matters.

\paragraph{Streams.}
Next to the EEG, the dataset records gaze, head motion and a head-mounted scene
video. All signals are resampled to $64$\,Hz and standardised per channel over
each trial. This gives four time-series streams --- EEG ($32$ channels), gaze
($6$), head motion ($6$) and optical-flow summaries of the scene video ($4$) ---
and, for each talker, the speech envelope. We add a fifth stream that the
released benchmark does not use: a \emph{frozen} V-JEPA\,2 ViT-L
\citep{assran2025vjepa2} embedding of the part of the scene video around the gaze
point, one $1024$-d vector per window, never trained.

\paragraph{Examples.}
Signals are cut into windows of $W$ seconds with hop $H$; unless stated,
$W\!=\!10$\,s and $H\!=\!5$\,s, which gives $T\!=\!640$ samples and $7{,}990$
windows. One example holds the streams $x^m\!\in\!\mathbb{R}^{T\times C_m}$; $K$
candidate envelopes, exactly one of them correct; the index $y$ of the correct
one; the attended loudspeaker; the permutation $\pi$ that records which
loudspeaker is in which candidate slot; and the listener's identity. The
candidates are put in a random order in every window, so their position says
nothing. At test time that order is fixed by the window's index, so it is the
same for every model we compare.

\paragraph{Two details of the candidate sets that matter.}
First, same-talker negatives must be drawn \emph{at random} from windows of the
same trial that do not overlap the target window. Taking the window furthest from
the target seems natural, but it pushes the negatives toward the start and end of
the trial, and the audio probe then learns the position: it reached $0.60$ on a
task with a chance of $0.50$. Second, a negative must never overlap the target
window; when a trial is too short to provide enough non-overlapping windows, we
lower $K$ rather than allow overlap. For per-trial decisions, the training crops
are drawn at random positions and the candidates are matched separately in each
crop, and the same-talker negatives are random non-overlapping crops of the same
trial.